\documentclass{article}
\usepackage[margin=3cm]{geometry} 
\usepackage[numbers,sort&compress,square]{natbib}
\usepackage{xcolor}
\definecolor{red}{RGB}{200,50,50}

\usepackage{mathtools} %
\usepackage{booktabs} %
\usepackage{tikz} %
\usetikzlibrary{shapes.geometric, arrows.meta, positioning, calc, fit}
\usepackage{sidecap}
\definecolor{darkgreen}{rgb}{0.0, 0.5, 0.0}
\usepackage{amsthm}
\usepackage{amsmath}
\usepackage{thmtools}
\usepackage{thm-restate}
\usepackage{amssymb}
\usepackage{bm}
\usepackage{clipboard}
\usepackage{svg}
\usepackage{mathtools}
\mathtoolsset{centercolon}
\usepackage{parskip}
\usepackage{graphicx}
\usepackage{multirow}
\usepackage{subcaption}
\usepackage{comment}
\usepackage{paralist, tabularx}
\usepackage{bbm}
\usepackage{changepage}
\usepackage{tikz}
\usepackage{verbatim}
\usepackage{algorithm}
\usepackage{algpseudocode}
\usepackage{hyperref}       %
\usepackage{cleveref}

\newif\ifsqueeze
\squeezefalse

\newcommand{\pa}{\mathbf{pa}}
\newcommand{\PA}{\mathbf{PA}}
\newcommand{\DE}{\mathbf{DE}}

\newcommand{\cov}{\text{cov}}
\newcommand{\var}{\text{var}}

\newcommand{\indep}{\mathrel{\text{\scalebox{1.07}{$\perp\mkern-10mu\perp$}}}}
\makeatletter
\newcommand*{\centernot}{%
  \mathpalette\@centernot
}
\def\@centernot#1#2{%
  \mathrel{%
    \rlap{%
      \settowidth\dimen@{$\m@th#1{#2}$}%
      \kern.5\dimen@
      \settowidth\dimen@{$\m@th#1=$}%
      \kern-.5\dimen@
      $\m@th#1\not$%
    }%
    {#2}%
  }%
}
\makeatother

\theoremstyle{definition}

\newtheorem{defi/}{Definition}[section]
\newenvironment{definition}
{%
	\pushQED{\qed}\begin{defi/}}
	{\popQED\end{defi/}}

\newtheorem{exam/}[defi/]{Example}
\newenvironment{example}
{%
	\pushQED{\qed}\begin{exam/}}
	{\popQED\end{exam/}}

\newtheorem{re/}[defi/]{Remark}

\newtheorem{set/}[defi/]{Setting}
\newenvironment{setting}
{%
	\pushQED{\qed}\begin{set/}}
	{\popQED\end{set/}}

 \newtheorem{tnp/}[defi/]{Theorem}

 \newtheorem{pnp/}[defi/]{Proposition}
 \newenvironment{propositionnoproof}
{%
	\pushQED{\qed}\begin{pnp/}}
	{\popQED\end{pnp/}}

  \newtheorem{coro/}[defi/]{Corollary}
 \newenvironment{corollary}
{%
	\pushQED{\qed}\begin{coro/}}
	{\popQED\end{coro/}}
	
\newtheorem{con/}[defi/]{Conjecture}

\newtheorem{notation/}[defi/]{Notation}

\def \rF {\text{\boldmath$F$}}

\def \rH {\text{\boldmath$H$}}

\def \rM {\text{\boldmath$M$}}

\def \rO {\text{\boldmath$O$}}

\def \rX {\text{\boldmath$X$}}

 \def \calC {\mathcal C}
 \def \calD {\mathcal D}

 \def \calG {\mathcal G}

 \def \calN {\mathcal N}

\usetikzlibrary{arrows.meta}
\tikzset{>=latex}

\tikzset{
    *|/.style={
        to path={
            (perpendicular cs: horizontal line through={(\tikztostart)},
                                 vertical line through={(\tikztotarget)})
            -- (\tikztotarget) \tikztonodes
        }
    }
}

\colorlet{DarkRed}{red!50!black}

\definecolor{linkcolor}{HTML}{338040}
\definecolor{citecolor}{HTML}{660022}
\definecolor{urlcolor}{HTML}{006666}
\hypersetup{
    colorlinks,
    linkcolor={linkcolor},
    citecolor={citecolor},
    urlcolor={urlcolor}
}

\author{
   \small Frederik Hytting Jørgensen\\
   \small Dept.\ of Mathematical Sciences\\
   \small University of Copenhagen\\
   \small Copenhagen, Denmark\\
   \small \texttt{frederik.hytting@math.ku.dk}
   \and
    \small Sebastian Weichwald\\
   \small Dept.\ of Mathematical Sciences\\
   \small University of Copenhagen\\
   \small Copenhagen, Denmark\\
   \small \texttt{sweichwald@math.ku.dk}
    \and
    \small Jonas Peters\footnote{Parts of this work were done while JP was at the University of Copenhagen}\\
  \small  Dept.\ of Mathematics\\
   \small ETH Zurich\\
    \small Zurich, Switzerland\\
   \small \texttt{jonas.peters@stat.math.ethz.ch}}

\title{Unfair Utility Functions\\ and First Steps Towards Improving Them}
\begin{document}
\maketitle

\begin{abstract}

{
Many fairness criteria constrain the policy or choice of predictors, which can have unwanted
consequences, in particular, when optimizing the policy under such constraints. Here, we instead suggest that fairness can be directly analyzed as a property of the utility function. Instead of 
imposing fairness constraints on the policy, we suggest to simply maximize a utility function satisfying certain fairness properties.  
Concretely, we define value of information fairness, which prescribes that there must not be an incentive to 
infer the protected attribute. This principle suggests
modifying utility functions such that they satisfy value of information fairness. We describe
how such modifications can be achieved and discuss consequences for the corresponding optimal policies. We apply
our framework to thought experiments and the COMPAS data, demonstrating that focusing on utility functions  sometimes provides answers that better align with intuitive judgments about what is fair.  Moreover, we are not aware of any intuitively fair policy that violates value of information fairness; and when we find that value of information
fairness recommends an intuitively unfair policy, no realizable policy is intuitively
fair.}

\end{abstract}

\medskip
\noindent\textbf{Keywords:} algorithmic fairness, utility functions, value of information, causal models

\medskip
\noindent\textbf{MSC 2020:} 62D20, 62A01

\section{Introduction}
A challenge
in algorithmic fairness is to formalize the notion of fairness.
Often, one attribute $S$ is considered protected (also called sensitive)
and a quantity $Y$ is to be predicted as $\widehat{Y}$ from some covariates $\rX$.
Many   
criteria for 
fairness
correspond to 
constraints on the joint distribution of $(S,X,Y,\widehat{Y})$
that can often be phrased as (conditional)
independence statements
or take the causal structure of the problem into account 
(see,
e.g.,
\citep{Verma2018,Nilforoshan2022,Barocas2019} for an overview; {for an introduction for practitioners, see \citep{leslie2024ai}}).
{ In this work, we propose an alternative to existing fairness approaches. We consider situations where an agent aims to optimize a policy}
as to maximize a known utility function.
In such scenarios, 
unwanted discrimination may occur
if
the utility function itself is unfair. 
 We formalize a constraint on the utility using value of information (VoI) 
\citep{Howard1966,jackson2022value} and call utilities that satisfy the constraint 
VoI-fair.\footnote{VoI is sometimes used to refer to the value
of additional observations \citep{raiffa1961applied}. In this work, it refers to the value of additional covariates.}
While it is not uncommon for works on fairness to include a utility function in the framework \citep{dwork2012fairness, hebert2018multicalibration,singh2021fairness,Nilforoshan2022,dehdashtian2024utility}, we are not aware of any existing works characterizing fairness as a property of the utility function.\footnote{Considerations of utilities are common in social choice theory \citep{sep-social-choice, fleurbaey2009beyond,heidari2018fairness}, while algorithmic fairness focuses mostly on the policies.}
We discuss examples of {VoI-}unfair utility functions 
that 
would lead to 
unwanted discrimination %
if used
(\Cref{ex: parental status}, \Cref{ex: doctor VoI fair}, \Cref{app: college admission}, and \Cref{ex: compas}). 
We provide a graphical criterion for VoI-fairness for when the underlying causal structure is known.
We
discuss how VoI-unfair %
utility functions may be improved to obtain %
VoI-fair utility functions that are close to the original utility function and apply 
these concepts in
experiments, including on the COMPAS data.
In general, the proposed utility-centric approach yields different 
implications than existing concepts of fairness and solves shortcomings of, for example, counterfactual fairness \citep{Kusner2017}, equalized odds \citep{Hardt2016}, and path-specific fairness \citep{Nabi2018,Chiappa2019,Kilbertus2017}. %
We explain these shortcomings and how they are dealt with by VoI-fairness in Examples \ref{ex: parental status}, \ref{ex: CF}, \ref{ex: EO}, and Appendix \ref{app: consequences} and \ref{app: path-specific}. 
\Cref{fig: decision procedure} in \Cref{sec: future work} provides a flow chart outlining a structured method for handling questions of fairness, derived from the insights of our work.

\subsection{Motivation and methodology}
`Fairness through unawareness', that is,
not using the protected attribute as decision input, 
fails because there may be other decision inputs available that are highly predictive of the protected attribute, thereby allowing proxy discrimination. 
We propose to obtain fairness
by using a utility function (also 
called a loss function) such that there is no incentive to infer the protected attribute. 
We then do not need 
to 
worry about the algorithm using other decision inputs as proxies for the protected attribute because the algorithm has no incentive to infer the protected attribute. 

{We formalize this idea using the concept of value of information.
Informally, a variable has value of information if it is possible to obtain a strictly higher utility by having access to this variable when making the decision (see Definition 3.2 for a formalization). Whether
the protected attribute has value of information depends on which other variables are already available. Therefore, care needs to be taken when defining VoI-fairness using VoI: }
(1), 
the protected attribute may have no value of information relative to the 
other decision inputs
because we can perfectly reconstruct 
it based on these; using these features may still allow for 
intuitively unfair policies. 
And (2), there
may also be some unwanted bias
in the data 
that we could 
adjust for by using the protected attribute, for example, a grade could be a biased estimate of true abilities, see \Cref{ex: grade};
if
the protected attribute has value of information 
relative to the biased data
and
can be used to correct for biases in the data, 
this
may be intuitively the right thing to do. 
Key to this work
(addressing both of the above points)
is that there exist features (which may be unobserved at decision time) relative to which the protected attribute should have no value of information; we
call such features essential features. Which features are to be considered essential is
a judgment made
using domain knowledge and ethical intuitions (this is similar to other
fairness frameworks, where one needs
to choose `resolving variables' \citep{Kilbertus2017} or `unfair paths' \citep{Chiappa2019}). For choosing the essential features, the following reasonings are often helpful: 
`An essential feature is a feature that would be legitimate to use to make a decision, even if the feature turned out to be 
substantially
correlated with the protected attribute;'
or
`The essential features are a set of features such that, were the essential features known at decision time, 
then it would be illegitimate to base the decision on the protected attribute in addition to the essential features.'
For example, 
when hiring medical staff, an essential feature may be the recovery rate of the applicant's patients, were the applicant to be hired.
In this case,
we propose to judge whether a utility is fair 
by checking whether
{ $S$ has value of information}
in a hypothetical situation where the recovery rate of the applicant's patients is observed before the hiring decision is made, see \Cref{ex: doctor VoI fair}. 
If such an incentive for using the protected attribute exists, we judge the utility function as unfair. 
Furthermore,
we propose to
not purposefully choose suboptimal policies under fair utilities.

Why do we consider changes to the utility rather than constraining the policy? 
First, whenever a fairness notion constrains the policy (and not the utility directly), one compromises the original utility. Thus, if we care about fairness and not just about finding optimal policies, then we are implicitly changing the utility function (traditional approaches may set the utility to minus infinity if the policy is disallowed). Our framework suggests to consider such changes explicitly.
Second, and more importantly, we believe that the framework presented in this paper does a better job of capturing what is intuitively fair in concrete examples than existing definitions of fairness.\footnote{
We call a policy intuitively fair if we perceive it to be morally appropriate within the 
idealizations of the thought experiment. We take intuitive fairness to be a property of policies; thus,
intuitive fairness depends on 
what
the usable decision inputs are and intuitive unfairness can occur because 
of lack of appropriate information even if the 
utility function is VoI-fair, see \Cref{sec:ODnotsubsetM}.}
When evaluating whether a particular definition of fairness is appropriate, human intuition is the ultimate arbiter (see also \citep{Nabi2018}). { The approach of refining general ethical principles (such as fairness definitions) in light of their consequences for concrete
example is sometimes explicitly endorsed as “the method of reflective equilibrium” \citep{Knight2025}.}
If the result of applying a given fairness definition is a policy that is intuitively less fair than another policy, then that is evidence against that definition of fairness. 
It is known that many fairness criteria can suggest intuitively unfair policies if applied, see Examples \ref{ex: parental status}, \ref{ex: CF}, \ref{ex: EO}, and Appendix \ref{app: consequences} and \ref{app: path-specific}. 
Our shift in focus 
provides better answers in scenarios where other fairness notions fail. We do not know of any examples where our approach recommends using an intuitively unfair policy, but an alternative existing fairness notion would pick an intuitively fair policy, see \Cref{sec:ODnotsubsetM}. 
Furthermore, we do not know of any examples of policies that are intuitively fair but disallowed by the framework presented in this article.  
\begin{example}[\textbf{Predicting parental leave at hiring}]\label{ex: parental status}
\begin{figure}
    \centering
   \begin{tikzpicture}[scale=1,thick,main/.style={draw,rectangle,anchor=west},every node/.style={scale=1}]
	\node[main,fill=cyan] (D) at(.55,0) {$D$};
	\node[main,fill=yellow] (U) at(2,0) {$U$};
	\draw[->] (D) to (U);	
	\node[main] at(-2.3,1)  (b1) {Qualifications};
	\node[main,align=center] at(-2.3,0)  (b2) {Application\\[-.25em]material, $A$};
	\node[main] at(-2.3,-1)  (b3) {Parental leave, $S$};
	\node[main,align=center] at(-4.5,0) (h) {Unknown\\[-.25em]factors};
	\draw[->] (b1.south) to[*|] (b2);
	\draw[->] (h) to (b1.west);
	\draw[->] (h) to (b2.west);
	\draw[->] (h) to (b3.west);
	\draw[->,dashed,DarkRed] (b2) to (D);
	\node[main] at(0.5,1) (p1) {Quality of work, $Q$};
	\node[main] at(1,-1) (p2) {Work hours, $W$};
	\draw[->] (p1.south) to[*|] (U.north);
	\draw[->] (p2.north) to[*|] (U.south);
	
	\draw[->] (b1) to (p1);
	\draw[->] (b3) to (p2);

        \node [draw, rectangle, rounded corners, thick, darkgreen, fit=(p1)] {};
        \node [draw, rectangle, rounded corners, thick, DarkRed, fit=(b2)] {};

        \node [draw, circle, thick, DarkRed, align=left,text=black] at(-4,-1) (od) {};
        \node [right=0.05cm of od] {$\rO^D$};
        \node [draw, circle,  thick, darkgreen, align=left,text=black] at(-4,1) (F) {};
        \node [right=0.05cm of F] {$\rF$};  
\end{tikzpicture}
    \caption{
The graph contains utility node $U$ and decision node $D$. 
For making the decision
only one feature is
available, namely application material. We call 
this feature a usable decision input
and denote it by
$\rO^D=\{A\}$. 
In this made up example, 
the graph admits value of information for parental leave relative to the essential feature $\rF = \{Q\}$; that is, we could obtain a higher expected utility using both quality of work and parental leave rather than just quality of work (\Cref{def:voi} and \Cref{prop: graphical}).
If parental leave
has VoI relative to $\{Q\}$, we say that $U$ is not $\{Q\}$-VoI-fair (\Cref{def: VoI-fairness}). 
\Cref{app: visual} provides further visual explanation of these concepts.}
\label{fig: parental status}
\end{figure}
Consider making a hiring decision based on application material and assume that the situation is as depicted in
\Cref{fig: parental status}. Quality of work and work hours cannot be used
by the policy since the values of these variables are only
realized after the decision.
Whether an applicant will take
parental leave may be partly predictable based on application material, for example, through gender and age. For this example, we assume that parental leave is the protected attribute.

Assume that we use the number of satisfied customers as a utility,
which is a function of the 
applicant's future
quality of work $Q$, work hours $W$,
and the hiring decision $D\in \{0,1\}$,
for example, $\upsilon(D,Q,W):=D\cdot Q \cdot W$. Arguably, making a hiring decision to optimize this utility function is incompatible with EU law because parental leave necessarily affects working hours.\footnote{``Member States shall take the necessary measures to prohibit less favorable treatment of workers on the ground that they have applied for, or have taken, leave provided for in Articles 4 [Paternity leave], 5 [Parental leave] and 6 [Carers' leave] ...'' Directive (EU) 2019/1158.}
One could argue that if $Q$ were available at decision time, then the utility function should not incentivize inferring parental leave. 
Since work hours $W$ enters the utility and is a
(noisy)
function of parental leave $S$,
it may be possible to obtain a higher expected utility using both $Q$ and $S$ rather than just $Q$. 
So even in a hypothetical situation where we could use quality of work as a decision input, there may still be an incentive to infer parental leave.  
If this is the case, we say the utility function is not $\{Q\}$-VoI-fair, see \Cref{def: VoI-fairness}.

If, however, we changed 
the utility function to directly take parental leave into account in a way such that there would be no benefit (in terms of utility) in having parental leave as decision input in addition to quality of work, we say that the modified utility function is $\{Q\}$-VoI-fair. This could be accomplished by considering `parental leave-adjusted working hours', that is, some quantity that renders parental leave useless as a decision input in addition to quality of work.{ More formally,
the utility function  $\widetilde{\upsilon}(D,Q,W,S)=D\cdot Q \cdot (W-\mathbb{E}(W\mid S))$ is $\{Q\}$-VoI fair.}

Finally,  by assumption, there is no causal effect from
parental leave to application material, so all policies based on application material will satisfy causal fairness definitions like counterfactual fairness and interventional fairness. This includes policies that try to infer parental leave from the application material, which is intuitively not the right thing to do (and possibly illegal). In \Cref{app: visual}, we provide a visual overview of the main ideas of this work applied to this example.
\end{example}

\subsection{Structure of this work}The remainder of this work is structured as follows. 
In \Cref{sec: setting}, we formally introduce the setting. \Cref{sec:voimain} contains the main definition of VoI-fairness and discussion of various properties. 
In \Cref{sec: improve utility}, 
we show how utility functions that are not VoI-fair could be modified to become VoI-fair. \Cref{sec: other workds} contains comparisons with existing notions, and 
\Cref{sec: experiments} contains experiments on simulated and real data.
{  Throughout the paper, we consider examples with known causal structure to aid intuition. This does not imply that VoI-fairness can only be applied when we have causal knowledge, which we show in \Cref{sec: experiments}. }

\Cref{fig: decision procedure} in \Cref{sec: future work} provides a flow chart outlining a structured method for handling questions of fairness, derived from the insights of our work.
\Cref{app: table examples} contains an overview of all examples in this paper.
All code is provided at \url{https://github.com/FrederikHJ/Unfair-Utilities}.

\section{Setting and Notation}\label{sec: setting}  %
We now 
introduce the model class that we use throughout the paper. %
We 
will consider variables $\rX$, a decision $D$, and a utility $U$. 
Our setting is similar to a contextual bandit setup %
where we usually say `action' 
instead of `decision' 
and `reward' 
instead of `utility' 
\citep[see, e.g.,][]{Sutton1998}.
In this work, the utility is a function whose expectation is maximized w.r.t.\ a policy.%
Formally, we consider a structural causal model \citep{Pearl2009} with one node being  
the decision $D$ and one %
sink node being 
the utility $U$ \citep[see also][]{Everitt2021}.
We only consider interventions on the utility and decision nodes.
Similar graphical models have been discussed 
as influence diagram graphs and influence diagrams
 \citep{Nilsson2000, Merwijk2022,Dawid2002}. {  In this work, we assume that the structural assignment of the utility is known. This assumption is essentially vacuous. E.g., if the utility is graduation from college, we 
do not assume knowledge about the causes of college graduation but only
that we know that the utility is college graduation.}

\begin{setting}\label{setting}
   We have
   variables $(\rX,D,U)=(X_1,\dots , X_d ,D,U)$, where $D$ is the \emph{decision}
   and $U$ is the \emph{utility}. One of the features, denoted by\footnote{By slight abuse of notation, we sometimes regard vectors, such as $\rX$,
   as a set.} $S\in \rX$, is a \emph{protected attribute}. 
We assume that $\rX$, $U$, and $D$ take values in $\mathbb{R}^d$, $\mathbb{R}$, and a finite set $\mathcal{D}$, respectively. 
  Let
$\rO^D \subseteq \{X_1, \ldots, X_d\}$ and $\rO^U \subseteq \{X_1, \ldots X_d,D\}$ represent variables that can be used as input
for the decision and the utility, respectively, which, in particular, implies that they are observed.  
  We assume that $\rO^D\subseteq \rO^U$. Let 
  \ifsqueeze
  $\Upsilon:=\{\upsilon\mid \upsilon: \mathbb{R}^{\left|\rO^U\right|} \rightarrow \mathbb{R}\}$  
  \else
  $\Upsilon:=\{\upsilon\mid \upsilon: \mathbb{R}^{|\rO^U|} \rightarrow \mathbb{R}\}$  
  \fi
  denote the set of utility functions. 
  {  A particular utility may not functionally depend on all variables in $\rO^U$. Therefore, we define $\PA^U$ (the \emph{parents of $U$}) as the subset of $\rO^U$ on which the utility functionally depends. Formally, for $\upsilon\in \Upsilon$,} 
  let $\PA^U\subseteq \rO^U$ 
  be 
  the unique smallest 
  subset of $\rO^U$ for which there exists a function 
  \ifsqueeze
  $\upsilon^*: \mathbb{R}^{|\PA^U|}\to \mathbb{R}$ 
  \else
  $\upsilon^*: \mathbb{R}^{\left|\PA^U\right|}\to \mathbb{R}$ 
  \fi
  such that $\upsilon^*(\pa^U)=\upsilon(\bm{o}^U)$ for all 
    \ifsqueeze
$\bm{o}^U\in \mathbb{R}^{|\rO^U|}$, 
\else
$\bm{o}^U\in \mathbb{R}^{\left|\rO^U\right|}$, 
\fi
  where $\pa^U$ is the coordinate projection of $\bm{o}^U$ onto $\PA^U$. 
  We can then regard $\upsilon$ as a function 
      \ifsqueeze
$\mathbb{R}^{|\PA^U|}\to \mathbb{R}$.
\else
$\mathbb{R}^{\left|\PA^U\right|}\to \mathbb{R}$.
\fi

We further consider measurable 
structural assignments $f_{\rX}=(f_1,\dots f_d)$ and jointly independent noise variables %
 $(\varepsilon_1,\dots, \varepsilon_d)$ with distribution $ \mathbb{P}_{\varepsilon_{\rX}}=\mathbb{P}_1\otimes \ldots  \otimes \mathbb{P}_d$.
 Each variable $X_i\in \rX$ is given by  $X_i:=f_{i}(\PA^i,\varepsilon_i)$
 for some $\PA^i\subseteq \rX\cup \{D\}$.
 Within each example that we discuss in this paper, we 
 consider the background model $\mathcal{C}:=(f_{\rX},\mathbb{P}_{\varepsilon_{\rX}},\rO^D,\rO^U,\calD)$ as fixed.
 Given $\calC$ and utility function $\upsilon$, we let $\mathcal{G}$ be the graph
 over $\rX$, $D$, and $U$
 induced by $f_{\rX}, \rO^D$, and $\PA^U$. {  For a variable $Y$, we denote the parents and descendants of $Y$ in $\mathcal{G}$ by $\PA^Y$ and $\DE^Y$, respectively.}
  We assume that 
  $\calC$ and  $\upsilon$ are such that
  $\calG$ is a DAG, in which $U$ is a descendant of $D$, and $U$ has no children 
  (this is called a causal influence diagram in \citep{Everitt2021}). {  We also assume that $S$ is a non-descendant of $D$.}
  {We indicate $\rO^D$ by red dashed arrows, see Figure~\ref{fig: parental status}, for example.} %
Let $\Delta(\calD)$ denote the set of probability distributions on
  $\calD$, and 
  for $\rM\subseteq \rX \backslash \DE^D$, let
  $\Pi^{\rM}$
  denote the set of policies taking $\rM$ as input.\footnote{
Formally, 
  $\Pi^{\rM}:=\{\pi: \mathbb{R}^{|\rX|}\to \Delta(\calD)
  \mid \exists \pi^{\rM}: \mathbb{R}^{\left|\rM \right|}  \rightarrow
  \Delta(\calD) \text{ s.t.\ } \forall x \; \pi(x) \equiv \pi^{\rM}(x^{\rM})\}$.
 By slight abuse of notation, for such $\pi$, we write expressions such as 
  $\pi(\rM)$. 
  We require that $\upsilon$ depends non-trivially on all variables in $\PA^U$,
  while $\pi \in \Pi^{\rM}$ may trivially depend
  on variables in $\rM$. }  If there exists a  $d\in \calD$ such that $\pi(\rM)(d)=1$, 
  we may write $\pi(\rM)=d$.
 Even though we have introduced $\rO^D$ as the input variables for $D$, we do not require that $\rM \subseteq \rO^D$; that is, we sometimes consider hypothetical situations where the decision is a function of variables that an actual decision could not be a function of. In the graph, we let $\rO^D$ be the parents of $D$. 
  We call $\rO^D$ the \emph{usable decision inputs}. 
 If $\pi \in \Pi^{\rO^D}$,
  we say that $\pi$ is \emph{realizable}.
    In summary, we consider models 
   containing a background model $\calC$, 
   a policy $\pi\in \Pi^{\rM}$ for hypothetical decision inputs $\rM\subseteq \rX \backslash \DE^D$, and a utility function $\upsilon\in \Upsilon$. 
  The background model $\mathcal{C}$ induces a class of structural causal models (SCMs). Choosing $\upsilon \in \Upsilon$, $\rM \subseteq \rX \backslash \DE^D$, and $\pi \in\Pi^{\rM}$ then fixes a unique SCM 
  and a unique distribution over $(\rX,D,U)$ %
  \citep{Pearl2009, Bongers2021}.
  This distribution can be described by the following sampling procedure: 
    
    \centerline{
        \begin{tabular}{ll}
(1) Sample $\rX \backslash \DE^D$.
 & 
   (2) Sample %
   $D\sim\pi(\rM)$. %
 \\
 (3) Sample $\DE^D\backslash \{U\}$.
&    (4) Compute 
$U:= \upsilon(\PA^U)$.
\end{tabular}}

  The sampling is done independently for different observations. We say that a policy $\pi\in \Pi^{\rM}$ is \emph{optimal in $\Pi^{\rM}$} %
  if $\mathbb{E}_{\pi}(U)\geq \mathbb{E}_{\pi'}(U)$ for all $\pi'\in \Pi^{\rM}$.
\end{setting}

\section{Value of information fairness
} \label{sec:voimain}
\subsection{Value of information}
The concept of value of information 
 dates back to \citep{Howard1966}; see also \citep{Everitt2021}. %
 A variable $S$ has VoI relative to $\rM$ if one can obtain a strictly larger expected utility if 
 the policy $\pi$ is allowed to be a function of $S$ in addition to $\rM$ compared to when $\pi$ is only a function of $\rM$.
 More formally, we have the following definition.
 \begin{definition}[\textbf{Value of information, VoI}]
 \label{def:voi}
A variable $S\in \rX \backslash \DE^D$ 
\textit{ has VoI relative to $\rM$ (under utility $U$)}  if
$$\underset{\pi\in \Pi^{\rM \cup \{S\}}}{\max}\mathbb{E}_{\pi}(U)>
\underset{\pi\in \Pi^{\rM}}{\max}\mathbb{E}_{\pi}(U).$$
 We further say that a DAG $\calG$ 
 over $(\rX, D, U)$,
where $U$ is a descendant of $D$ and $U$ has no children, \emph{admits value of information 
for $S$ relative to $\rM\subseteq \rX \backslash \DE^D$} 
if there exists a background model $\calC$ and utility function $\upsilon$ such that $S$ has VoI relative to $\rM$ and the induced graph equals $\calG$.
\end{definition}
Whether $S$ has VoI relative to $\rM$ does not depend on $\rO^D$. The concept of VoI has already proven useful for analyzing algorithmic fairness; see \citep{Ashurst2022} for conditions under which algorithms may amplify disparities. We use it, instead, to analyze unfair utility functions.
\subsection{Value of information fairness}

We introduce two concepts of fairness: (1)
VoI-fair utility functions and, (2), VoI-fair policies.
In essence,
a policy is 
VoI-fair if it optimizes a VoI-fair utility function. 
{  The essential features are a subset of $\rX \backslash (\DE^D\cup \{S\})$. Which subset we choose as the essential features depends on the context, see Section 1.1.} 
Assume that we are given a
protected attribute $S\in \rX \backslash \DE^D$ and a set of essential
features $\rF\subseteq \rX\backslash (\DE^D \cup \{S\})$. {(In some situations, even $S$ might be an essential feature. Examples may
include considering parental status an essential feature for receiving child benefits. In such cases, VoI-fairness is trivially satisfied.)}
If there is an incentive to infer $S$ 
in the situation where 
the essential features are known, we say that a utility function violates 
value of information fairness.
\begin{definition}[\textbf{Value of information fairness}]\label{def: VoI-fairness}

Let background model $\calC$ and a set 
$\rF \subseteq \rX \backslash (\DE^D\cup \{S\})$
of \emph{essential features} 
be given.
We say that a utility function 
$U:=\upsilon(\rO^U)$
satisfies \emph{$\rF$-VoI-fairness} if $S$ does not have VoI relative to $\rF$ under utility $U$.
\end{definition}
This definition does not depend on 
the policy being used. It does not 
depend on the
usable decision inputs $\rO^D$, either, that is,
{  whether or not a utility function satisfies VoI-fairness does not hinge on what variables are available for making decisions (however, $\rO^D$ 
is relevant when assessing if a policy is VoI-fair, see Definition 3.3).
}

Given an 
$\rF$-VoI-fair utility function, 
we 
now
define an 
$\rF$-VoI-fair policy 
as an optimal realizable
policy that is a function only of {  `useful variables'}, that is,
$\pi\in \Pi^{\rH}$ for a subset $\rH\subseteq \rO^D$ where each variable $X\in \rH$ has VoI relative to $\rH\backslash \{X\}$. 
This requirement ensures
 that an $\rF$-VoI-fair policy 
 is not arbitrarily discriminating based on irrelevant features (see \Cref{app: VoI-fair policy} for details). 
 More precisely, we have the following definition.
\begin{definition} [\textbf{VoI-fair policy}]\label{def: VoI-fair policy}
Let background model $\calC$, utility function $\upsilon$,
and essential features $\rF\subseteq \rX \backslash (\DE^D\cup \{S\})$ be given.
 Assume that $\upsilon$ is an $\rF$-VoI-fair utility function. 
We say that $\pi\in \Pi^{\rO^D}$ is a \emph{$\rF$-VoI-fair policy} if it is an optimal policy in $\Pi^{\rO^D}$, and there exists a subset $\rH \subseteq \rO^D$ such that $\pi\in \Pi^{\rH}\subseteq \Pi^{\rO^D}$ and each variable $X\in \rH$ has VoI relative to $\rH \backslash \{X\}$.
\end{definition}
Whether a policy satisfies the above definition does not only depend on essential features $\rF$ but also on usable decision inputs $\rO^D$.
Sections~\ref{sec: graphical}--\ref{sec:ODnotsubsetM} discuss various properties of VoI-fair policies. E.g.,  
a policy being $\rF$-VoI-fair does not imply that it is necessarily intuitively fair, see \Cref{sec:ODnotsubsetM}. 
As far as we know, this occurs only in situations where no alternative fairness notion finds an
intuitively fair policy.
Based on these insights, we propose the
decision diagram in \Cref{fig: decision procedure}.

\subsection{Graphical criteria for value of information fairness}\label{sec: graphical}
The following proposition establishes a graphical criterion for the concept of admitting VoI.

\begin{propositionnoproof}[\textbf{Graphical criterion for admitting VoI}]
\label{prop: graphical}
\Copy{propgraphical1}{A DAG $\mathcal{G}$ over 
$(\rX, D, U)$, where $U$ is a descendant of $D$ and $U$ has no
children, admits VoI for $S\in \rX \backslash \DE^D$ relative to $\rM\subseteq \rX \backslash (\DE^D\cup \{S\})$ if and only if}
\begin{equation} \label{eq:invpr}
 {S  \not \perp_{\calG_{\PA^D:=\rM}} U \mid \rM.}
\end{equation}
\Copy{propgraphical2}{Here,
$\perp_{\calG_{\PA^D:=\rM}}$
denotes $d$-separation in the DAG $\mathcal{G}_{\PA^D:=\rM}$
obtained from $\mathcal{G}$ by modifying the parents of $D$ to be $\rM$.}
\end{propositionnoproof}
We provide a proof in \Cref{app: proof graphical}. The statement follows from %
Theorem 9 in \citep{Everitt2021}.
A related non-invariance property is discussed in \citep{Saengkyongam2023}. It is 
non-invariance property that is  similar to~\eqref{eq:invpr} with $S$ being the environment.

The following result provides a graphical criterion for VoI-fairness; it follows from \Cref{prop: graphical}. 
{  The graphical condition is a sufficient condition:
a utility $U$ can be $\{\rF\}$-VoI fair without the d-separation holding in the causal graph, see \Cref{ex: parental status} and \Cref{ex: doctor VoI fair}. Also, it is not necessary to know the causal graph to find a
VoI-fair utility function, see \Cref{sec: experiments}.}

\begin{propositionnoproof}[\textbf{Graphical criterion for VoI-fairness}]\label{prop: VoI-fairness graphical}
Let background model $\mathcal{C}$, essential features $\rF\subseteq \rX \backslash (\DE^D \cup \{S\})$, and utility function $\upsilon$ be given, and let $\mathcal{G}$ be the induced graph.~%
If
\ifsqueeze
    $S \perp_{\mathcal{G}_{\PA^D:=\rF}} U \ | \ \rF,$
\else
\begin{align*}
    S \perp_{\mathcal{G}_{\PA^D:=\rF}} U \ | \ \rF,
\end{align*}
\fi
then
$\upsilon$ is $\rF$-VoI-fair. 
\end{propositionnoproof}
{  Applying Proposition 3.5 to Example 1.1, see Figure 1, we can conclude that a utility that is only a function of quality of work would be $\{Q\}$-VoI-fair, since if there were no edge $W\to U$, then $S\perp_{\mathcal{G}_{\PA^{D}:=Q}} U\mid {Q}$. In general,} \Cref{prop: VoI-fairness graphical}  implies that VoI-fairness holds
if the utility is a function only of essential features 
and the decision.
\begin{corollary}\label{prop: H}
Let background model 
$\mathcal{C}$, essential features $\rF \subseteq \rX \backslash (\DE^D\cup \{S\})$, and utility function $\upsilon$ be given. 
If $\PA^U \subseteq \rF\cup \{D\}$, then $U$ is $\rF$-VoI-fair. 
\end{corollary}

\subsection{A VoI-fair policy may need to be a non-constant function of S} \label{sec:pidependsonS}
Even when $S$ does not have VoI relative to $\rF$, it is still possible that $S$ has VoI in the context where the decisions are actually being made, that is, relative to $\rO^D\backslash \{S\}$. 
The corresponding $\rF$-VoI-fair policy may then be a non-constant function of $S$, as the following example illustrates.
\begin{example}[\textbf{College admission based on grade and $S$}]\label{ex: grade}
We consider the case of college admission based on high school grades and $S$, that is, $\rO^D=\{S,\text{Grade}\}$, see \Cref{fig: grade}. $D\in \{0,1\}$ is the admission decision.
	\begin{figure}
 \centering
		\begin{tikzpicture}[node distance={20mm},thick,xscale=1.0,main/.style={draw,rectangle}]
		\node[main,fill=none] (1) at(0,0) {Grade};
		\node[main,fill=cyan] (2) at (1.5,0){$D$};
		\node[main,fill=yellow] (3) at (3,0){$U$};
		\node[main,fill=none] (4) at(0.75,1) {$S$};
		\node[main,fill=none] (5) at (-2,0 ){Abilities};
		\draw[->] (5) to (1);
		\draw[->,dashed,DarkRed] (1) to (2);
		\draw[->] (2) to (3);
		\draw[->] (4) to (1);
		\draw[->,dashed,DarkRed] (4) to (2);
		\draw[->] (5.south east) to [out=-18,in=200,looseness=1](3);
		\end{tikzpicture}
		\caption{%
  This figure depicts the graph $\calG$ induced by background model $\mathcal{C}$ and utility function $\upsilon$ from \Cref{ex: grade} with $\rO^D=\{\text{Grade}, S\}$. 
		$\mathcal{G}$ does not admit VoI for $S$ relative to $\{\text{Abilities}\}$, so $U$ is $\{\text{Abilities}\}$-VoI-fair, but $\mathcal{G}$  does admit VoI for $S$ relative to $\{\text{Grade}\}$. Indeed, an $\{\text{Abilities}\}$-VoI-fair policy will use $S$ 
		to obtain a better estimate of abilities.
  } 
        \label{fig: grade}
	\end{figure} 
Here, we consider abilities an essential feature
$\rF=\{\text{Abilities}\}$. We have that $S\bot_{\mathcal{G}_{\PA^D:=\{\text{Abilities}\}}} U \ | \ \left\{\text{Abilities}\right\}$, so any $U$ that is a function only of abilities is $\{\text{Abilities}\}$-VoI-fair by \Cref{prop: VoI-fairness graphical}. 
But $\mathcal{G}$ does admit VoI for $S$ relative to $\rO^D \backslash \{S\}=\{\text{Grade}\}$. 

Let us assume the following assignments
(using $A$ as shorthand for abilities and $G$ for grade):
\begin{align*}
    &S:= \varepsilon_S\sim \text{Unif}\left(\{-1,1\}\right), \qquad A:= \varepsilon_A\sim \mathcal{N}(0,1), \\
    &G:=A+\alpha S, \qquad U:=\upsilon(D,A)=\mathbbm{1}_{(D=1)}\cdot A, 
\end{align*}
where $\alpha\in \mathbb{R}\backslash \{0\}$. 
Here, $\alpha S$ is a term describing unwanted discrimination based on $S$. 
 The policy 
    $\pi(G,S)=\mathbbm{1}(G-\alpha S\geq 0)$ %
 is 
 $\{\text{Abilities}\}$-VoI-fair 
since 
it is optimal
in $\Pi^{\rO^D}$. 
The fact that $S$ has VoI relative to $\rO^D\backslash \{S\}$ does not imply that the utility function is not $\{\text{Abilities}\}$-VoI-fair. %
Indeed, if the policy did not use $S$, that is, if $\pi \in \Pi^{\{G\}}$, then the policy could not adjust for the discrimination $\alpha S$ in grade.  However, observing $S$ is not always sufficient to adjust for unwanted discrimination in the data, see \Cref{ex: entangled discrimination}. 
\end{example}

\subsection{A VoI-fair policy may be considered intuitively unfair}\label{sec:ODnotsubsetM}
In some situations,  
the data available for a realizable policy, $\rO^D$, may
be such that none of the realizable policies (including VoI-fair policies)
can be considered intuitively fair. We now provide an example of this type.
In \Cref{app: unfair}, we provide three more examples where we consider
unequal data quality, unavailability of $S$, and entangled unwanted discrimination in data. 
In all these examples, we regard it as unclear which realizable policy should be preferred over the VoI-fair policy.
Therefore, we consider VoI-fairness to be a necessary but not sufficient condition for intuitive fairness. And we suggest that a VoI-fair policy may be the best possible option given the available data.
\begin{example}[\textbf{Unfairness due to unavailability of relevant data}]\label{ex: hair}
    \begin{figure}
    \centering
    \begin{tikzpicture}[node distance={20mm},xscale=1.0,thick,main/.style={draw,rectangle},scale=0.5]
    \node[fill=none,style=main](S) at (-4,-2) {$S$};
    \node[fill=none,style=main](H) at (0.5,-2) {Hair length};
    \node[fill=none,style=main](P) at (0.5,-4) {Physical strength};
    \node[fill=cyan,style=main](D) at (4.5,-2) {$D$};
    \node[fill=yellow,style=main](U) at (8,-2) {$U$};
    \draw[->] (S) to (P.west);
    \draw[->] (S) to (H);
    \draw[->,dashed,DarkRed] (H) to (D);
    \draw[->] (D) to (U);
        \draw[->] (P.east) to (U);
    \end{tikzpicture}
    \caption{Graph of  \Cref{ex: hair}, where $\rO^D=\{\text{Hair length}\}$.
    If physical strength is an essential feature, a VoI-fair policy will be a policy that tries to infer physical strength based on hair length.
    Here, it seems impossible to make fair decisions. Even if it may seem repugnant to base decisions on hair length only, it is not clear that it is worse than a constant policy in $\Pi^{\emptyset}$. 
    }\label{fig: hair}
    \end{figure}
Assume that we are hiring a person to perform physically demanding labor, and for some reason, we only have information about the applicant's hair length, $\rO^D=\{\text{Hair length}\}$, see \Cref{fig: hair}. We assume 
the following structural assignments 
(using $P$ and $H$ as shorthand for physical strength and hair length, respectively):
\begin{align*}
    &S:=\varepsilon_S\sim \text{Unif}(\{-1,1\}), 
    \quad
    P:=\theta_S^{P}S+\varepsilon_{P},\\ 
     &H:=\theta_S^{H} S +\varepsilon_{H}, 
     \quad
     U:=\mathbbm{1}_{(D=1)}\cdot P, 
\end{align*}
where $\varepsilon_{P},\varepsilon_{H} \sim \mathcal{N}(0,1)$ 
and $\theta_S^H,\theta_S^P >0$.
Assume that physical strength 
is an essential feature, $\rF=\{P\}$. Then 
$\pi(H)=1(H>0)$
is a $\{P\}$-VoI-fair policy.
Intuitively, it seems unfair to base a decision on 
$H$; 
a person could, for example,
influence the hiring decision by getting a haircut before the interview.
In this example,
available data seems insufficient for choosing a fair policy. 
\end{example}

\section{Improving utility functions}\label{sec: improve utility}
In this section, for illustrative purposes, we first present a mathematically explicitly worked out example of a VoI-unfair utility and a possible adjustment. We then propose a general notion of VoI-fair utilities that `correspond' to the original utility.
\subsection{An unfair utility function and a possible improvement}\label{sec:unfairutility}
The following example shows an unfair utility function and suggests a way to improve 
it. 
\begin{example}[\textbf{Hiring medical staff}]\label{ex: doctor VoI fair}
    Consider the following variables from a hiring setting:
\begin{align*}
&S:=\varepsilon_S\sim \text{Unif}(\{-1,1\}) \qquad 
M':=\theta_S^{M'}S+\varepsilon_{M'},\\ 
&N':=\theta_S^{N'} S +\varepsilon_{N'} \qquad \qquad  \quad
M:=\theta_{M'}^M M'+\varepsilon_M,\\
&N:=\theta_{N'}^N N'+\varepsilon_N, \\
&U:=\upsilon(D,M,N)=\mathbbm{1}_{(D=1)}\cdot(\theta_N^U N+ \theta_M^U M),
\end{align*}
where $S$ is the protected attribute of applicant,
$M'$ an objective measure of medical qualifications,
$N'$ a measure of how much the interviewers like the applicant,
$M$ recovery rate of applicant's patients,
$N$ an evaluation by colleagues, and 
$U$ job performance evaluation after first year,
see \Cref{fig: med}. 
   \begin{figure}
		\centering
		\begin{tikzpicture}[node distance={20mm},thick,main/.style={draw,rectangle},scale=0.5]
		\node[fill=none,style=main](S) at (0,-2) {$S$};
		\node[fill=none,style=main](Mp) at (3,-.3) {$M'$};
		\node[fill=none,style=main](Np) at (3,-3.7) {$N'$};
		\node[fill=none,style=main](M) at (6,-.3) {$M$};
		\node[fill=none,style=main](N) at (6,-3.7) {$N$};
		\node[fill=cyan,style=main](D) at (4.5,-2) {$D$};
		\node[fill=yellow,style=main](U) at (9,-2) {$U$};
		\draw[->] (S) to (Mp);
		\draw[->] (S) to (Np);
		\draw[->,dashed,DarkRed] (Mp) to (D);
		\draw[->] (M) to (U);
		\draw[->,dashed,DarkRed] (Np) to (D);
		\draw[->] (N) to (U);
		\draw[->] (D) to (U);
		\draw[->,dashed,DarkRed] (S) to (D);
		\draw[->] (Mp) to (M);
		\draw[->] (Np) to (N);
		\end{tikzpicture}
        \caption{This figure depicts the graph induced by $\mathcal{C}$ and $\upsilon$ from \Cref{ex: doctor VoI fair}. The original utility function
        suggests
        optimizing for a quantity that contains a human bias, causing the utility function to not be $\{M\}$-VoI-fair.
        Using the concept of VoI-fairness (\Cref{def: VoI-fairness}), we can identify this bias and adjust for it.\label{fig: med}}
	\end{figure}
 We assume $\varepsilon_{M'},\varepsilon_{N'},\varepsilon_{M},\varepsilon_{N}\sim  \mathcal{N}(0,1)$ and 
{$\theta_S^{M'}, \theta_S^{N'}, \theta_{M'}^M, \theta_{N'}^N, \theta_M^U, \theta_N^U > 0$}.
    We assume that 
    $\rO^D=\{M',N',S\}$, $\calD=\{0,1\}$, and that the recovery rate is the only essential feature, that is, $\rF=\{M\}$. We assume that colleagues and interviewers have 
    an unwanted bias in their evaluations, corresponding to the term $\theta_S^{N'} S$. Because of this bias, $S$ has VoI relative to $\{M\}$, so $U$ is not $\{M\}$-VoI-fair.
    One possible way to modify $U$ is to instead consider
    \ifsqueeze
$\widetilde{U}:=\widetilde{\upsilon}(D,M,N,S)=     \mathbbm{1}(D=1) \cdot (\theta_{M'}^M\theta_M^U M'+\theta_{N'}^N\theta_N^U N'-\theta_S^{N'}\theta_{N'}^N\theta_N^U S)
$
\else     
\fi
\begin{align*}
    \widetilde{U}&:=\widetilde{\upsilon}(D,M,N,S)\\ &= \mathbbm{1}_{(D=1)} \cdot (\theta_M^U M+\theta_N^U N-\theta_S^{N'}\theta_{N'}^N\theta_N^US)\\
& = \mathbbm{1}_{(D=1)} \cdot (\theta_{M'}^M\theta_M^U M'+\theta_{N'}^N\theta_N^U N'-\theta_S^{N'}\theta_{N'}^N\theta_N^U S)
    \end{align*}
As
$\widetilde{U} = \mathbbm{1}(D=1) \cdot \left(\theta_{N}^U(\varepsilon_{N}+\theta_{N'}^N\varepsilon_{N'})+\theta_M^U M \right)$, 
$\widetilde{U}$ is $\{M\}$-VoI-fair
with corresponding $\{M\}$-VoI-fair policy
    \ifsqueeze
            $\widetilde{\pi}(M',N',S)=\mathbbm{1}(\theta_{M'}^M\theta_M^U M'+\theta_{N'}^N\theta_N^U N'-\theta_S^{N'}\theta_{N'}^N\theta_N^U S \geq 0)$.
    \else
    \begin{align*}
            \widetilde{\pi}(M',N',S)=\mathbbm{1}_{\left(\theta_{M'}^M\theta_M^U M'+\theta_{N'}^N\theta_N^U N'-\theta_S^{N'}\theta_{N'}^N\theta_N^U S \geq 0\right)}.
    \end{align*}
    \fi
    The optimal policy in $\Pi^{\rO^D}$ under utility $U$ is
        $\pi(M',N',S)=\mathbbm{1}\left(\theta_{M'}^M\theta_M^U M'+\theta_{N'}^N\theta_N^U N' \geq 0\right)$.
    The {discrepancy} between the two policies, {the term} $-\theta_S^{N'}\theta_{N'}^N\theta_N^U S$, is a penalty/bonus given on the basis of $S$ to correct for the unwanted bias in evaluations by colleagues and interviewers. 
   {  The algorithmic nature of the hiring method, the quantifications of variables, and explicit knowledge of the utility's structural assignment make it, in principle, possible to adjust for biases by adjusting the utility function,  something that cannot be done in a non-algorithmic setting, see \Cref{ex: doctor VoI fair continued} for experiments related to this example.
}
    Other $\{M\}$-VoI-fair utility functions exist. Given an original utility function $\upsilon$, we discuss the notion of an $\rF$-VoI-fair utility function corresponding to $\upsilon$ in \Cref{sec: corresponding} (see \Cref{ex: doctor VoI fair continued} for 
    more on
    this example).

Under the original utility function $\upsilon$, even {  the policy  defined by 
$
D=1 :\Leftrightarrow \upsilon(D=1,M,N)>0   
$
(which we call the oracle policy),} that is, hiring exactly those people who would get a positive score 
under the original performance measure, would be intuitively unfair because the performance measure includes an unwanted human bias. 
Hiring exactly those who would get a positive score 
under the modified performance measure,
\begin{align} \label{eq:orpol2}
   D=1 :\Leftrightarrow \widetilde{\upsilon}(D=1,M,N,S)>0,
\end{align}
seems intuitively fair but is not realizable.
\end{example}
In \Cref{sec: VoI-fair policy is best}, we argue that a VoI-fair policy may be a realizable policy closest to the unrealizable oracle policy {  defined above}.
In \Cref{app: path-specific}, we investigate the consequences of applying path-specific fairness to \Cref{ex: doctor VoI fair}.
\subsection{Corresponding VoI-fair utility functions}\label{sec: corresponding}
In Section~\ref{sec:unfairutility}, 
we have 
seen an example of how  
an unfair utility function can be modified to become VoI-fair. 
Often, there are several 
$\rF$-VoI-fair utility functions
(for example, 
a constant utility function is always $\rF$-VoI-fair, for any $\rF\subseteq  \rX \backslash (\DE^D \cup S)$).
How do we decide which one to use?
We now introduce 
corresponding VoI-fair utility functions as one possibility.
   Given a utility function $\upsilon$ that is not $\rF$-VoI-fair, we aim to find an $\rF$-VoI-fair utility function $\widetilde{\upsilon}$ such that $\upsilon(\rO^U)$ and $\widetilde{\upsilon}(\rO^U)$ are similar under all policies. %
\begin{definition}[\textbf{Corresponding VoI-fair utility functions}]\label{def: appropriate}
    Let background model $\calC$, utility function $\upsilon$, and essential features $\rF\subseteq \rX \backslash (\DE^D\cup \{S\})$ be given. Let $\Upsilon_{\rF}\subseteq \Upsilon$ be the set of utility functions that are $\rF$-VoI-fair. We call a modified utility function $\widetilde{\upsilon}\in \Upsilon_{\rF}$ 
    satisfying
	\begin{equation*}
	\widetilde{\upsilon}\in \underset{\upsilon'\in \Upsilon_{\rF}}{\arg \min} \underset{\pi \in \Pi^{\rO^D}} {\sup}\mathbb{E}_{\pi} (\upsilon(\rO^U)-\upsilon'(\rO^U))^2
	\end{equation*}
 an \emph{$\upsilon$-corresponding $\rF$-VoI-fair utility function}.
\end{definition}
In practice, 
it is sometimes easier to find a $\upsilon$-corresponding $\rF$-VoI-fair utility function when we restrict the set of utility functions to a subset $\Upsilon^*\subseteq \Upsilon$. 
In this case, we say that $\widetilde{\upsilon}\in  \Upsilon^*_{\rF}\subseteq \Upsilon^*$ is the $\upsilon$-corresponding $\rF$-VoI-fair utility function in $\Upsilon^*$. It may occur that $\Upsilon^*$ does not contain an $\rF$-VoI-fair utility function, in which case one may instead try to find a utility function that minimizes VoI of $S$ relative to $\rF$ within $\Upsilon^*$. %
Which utility functions satisfy \Cref{def: appropriate} may depend on which features one considers essential. 
 A questionable choice for the set of essential features can lead to 
a $\upsilon$-corresponding $\rF$-VoI-fair utility function that seems unintuitive:
In \Cref{ex: hair depends on M} {  in \Cref{app: hair depends on M}} we illustrate how the VoI-fair policy in \Cref{ex: hair} depends on which features are considered essential.

\section{Relation to existing work}\label{sec: other workds}

Since the perspective taken in this article, namely considering fairness as a property of utility functions, is a unique perspective in the fairness literature, direct comparisons with existing notions is not straightforward. Yet, we believe that applying existing fairness notions in concrete examples leads to undesired consequences, which VoI-fairness avoids.

\subsection{Comparison with counterfactual fairness} \label{sec:counterfacfair}
Counterfactual fairness \citep{Kusner2017} is a promising approach in algorithmic fairness.
A policy
$\pi\in \Pi^{\rO^D}$ 
is called counterfactually fair if it
satisfies
    $\mathbb{P}_\pi(D_{S:=s}=d\mid \rO^D)=\mathbb{P}_\pi(D=d\mid \rO^D),$
for all settings $s$ of $S$ and $d$ of $D$;
here, $D_{S:=s}$ is the potential outcome of $D$ if $S$ is set to $s$.
The approach has been criticized for having shortcomings with respect to philosophy \citep{Hu2020,Kasirzadeh2021}, selection bias \citep{Fawkes2022}, and identification \citep{Yongkai2019}. In this section, we focus on the problem highlighted 
in \citep{Nilforoshan2022}, namely that under weak conditions, counterfactual fairness can only be satisfied by a policy 
that also is
in $\Pi^{\rO^D\backslash \DE^S}$.

\begin{example}[\textbf{VoI-fairness may solve shortcomings of counterfactual fairness}]\label{ex: CF}
Consider again the {setting} from \Cref{ex: grade} but with the following structural assignments 
       \begin{align*}
        &S \sim\text{Unif}(\{-1,1\}), \;  A\sim \mathcal{N}(0,1), \; U:=\mathbbm{1}_{(D=1)}\cdot A\\
	&G:=\mathbbm{1}_{(S=-1)}(A+\alpha S+\varepsilon_0) + \mathbbm{1}_{(S=1)}(A+\alpha S+\varepsilon_1),   
	\end{align*}
with $\begin{pmatrix}
    \varepsilon_0\\
    \varepsilon_1
\end{pmatrix}\sim \mathcal{N}\left(\begin{pmatrix}0\\0 \end{pmatrix}, \begin{pmatrix}
    1 & \sigma\\
    \sigma & 1 
\end{pmatrix}\right)$, $\sigma\in (-1,1)$ and $\alpha \in \mathbb{R}\backslash \{0\}$.  The setup is identical to \Cref{ex: grade}, except that the two different groups have distinct noise terms $\varepsilon_0,\varepsilon_1$ associated with them. A counterfactual fair, realizable policy $\pi\in \Pi^{\{G,S\}}$ must be a policy in $\Pi^\emptyset$ (this follows from Proposition F.8. in \citep{Nilforoshan2022}, see \Cref{app: CF}).
In this example, $\pi(G,S)=\mathbbm{1}(G-\alpha S\geq 0)$ 
 is an $\{\text{Abilities}\}$-VoI-fair policy. 
 \end{example}

\subsection{Path-specific causal fairness} In \Cref{app: path-specific}, we explain how path-specific counterfactual fairness has a similar problem as the one considered in \Cref{ex: CF}. More generally, causal fairness notions 
often  
suggest using intuitively unfair policies in
situations like the one presented in the motivating \Cref{ex: parental status} where there is no causal effect of the protected attribute on the decision. 

 \subsection{Comparison with equalized odds}\label{sec:eqodd}
In \Cref{ex: CF}, all policies satisfying the constraint implied by counterfactual fairness
seem intuitively unfair.
Generally, there may exist different policies satisfying a given fairness constraint, and some of these policies may be intuitively unfair.
Consider the example of enforcing demographic parity, that is, $D\indep S$: This constraint does not exclude that, for example, we pick the least qualified candidates 
from a disadvantaged group \citep{Hardt2016}. %
The following example illustrates that 
enforcing equalized odds may have unintended consequences, too.
\begin{example}[\textbf{Shortcomings of equalized odds}]\label{ex: EO}
    Consider again the example of college admission:
	$S:=\varepsilon_S\sim\text{Unif}(\{0,1\})$, 
 $A:=\mathbbm{1}(S=0)\sqrt{2}\varepsilon_{A_{0}} + \mathbbm{1}(S=1)\varepsilon_{A_1}$,
	$U:=\mathbbm{1}(D=1)\cdot \left(S-0.5\right)$,
with $\varepsilon_{A_0},\varepsilon_{A_1} \overset{iid}{\sim} \calN(0,1)$, and assume that the usable decision inputs are $\rO^D=\{A,S\}$, that is, we directly observe abilities. 
We assume that the utility function depends on $S$ only and not on $A$ because of blatant discrimination. We may try to fix this by imposing some fairness constraint like equalized odds \citep{Hardt2016}, that is, $D\indep S \mid A$.
The optimal policy in $\Pi^{\rO^D}$ that satisfies this constraint is $\widetilde{\pi}(S,A)=\mathbbm{1}(A\in [-c,c])$ for $c\approx 1.18$,
which is 
an intuitively unfair
policy, for example, because applicants with high medical qualifications $> c$ are not hired. 
The policy uses $A$ to infer $S$ and exploits this information to optimize the 
utility function. 
Here, 
the utility function is not $\{A\}$-VoI-fair. 
\Cref{app: consequences EO} shows 
that equalized odds may be intuitively inappropriate to impose even if the utility function is VoI-fair.
Therefore, enforcing equalized odds can be inappropriate regardless of whether the utility is VoI-fair or not.
Conditional statistical parity \citep{Corbett-Davies2017}, using $A$ as the `legitimate feature', is equivalent to equalized odds in this example. 
\end{example}

Applying fairness constraints like equalized odds is also known to result in decisions that are Pareto suboptimal from the perspective of balancing original utility and preferences for diversity \citep{Nilforoshan2022}, see also example \Cref{app: college admission} for a discussion of this point.

\section{Experiments} \label{sec: experiments}
In this section, we consider three experiments. In \Cref{ex: doctor VoI fair continued}, we explicitly calculate the $\upsilon$-corresponding VoI-fair utility and show how to recover it from observational data. In \Cref{app: college admission}, we show how  VoI-fairness can be applied in a setting with binary decisions. Finally, on the COMPAS dataset, we show how to apply VoI-fairness to a prediction task. In none of the examples, do
we make use of causal knowledge of the system. These examples illustrate how VoI-fairness can be applied both to the task of fair decision-making, where a specific decision has to be made (\Cref{ex: doctor VoI fair continued} and \Cref{app: college admission}) and to the task of making fair predictions (\Cref{ex: compas}) \citep{plevcko2024causal}. In \Cref{ex: compas}, we  illustrate statistical testing of a VoI-unfair utility function. 
\subsection{Synthetic experiments}

In this section we illustrate examples of applying the VoI-fairness framework
in
synthetic experiments. In \Cref{ex: doctor VoI fair continued} we 
show, in the context of \Cref{ex: doctor VoI fair}, how to empirically estimate an $\upsilon$-corresponding
$\rF$-VoI-fair utility function.
This example and the related appendices sketch how VoI-fairness may be applied in settings where we only have data from a fixed policy. In \Cref{app: college admission}, we illustrate how
the framework of VoI-fairness can be applied as a principled way to choose a utility offset $\omega S$ that incentivizes diversity. This example also 
illustrates the non-trivial task
of choosing essential features.

\subsubsection{Hiring medical staff}\label{ex: doctor VoI fair continued}
Here, we consider the setting described in \Cref{ex: doctor VoI fair}, where
$\rO^D=\{M',N',S\}$, $\calD=\{0,1\}$, and $\rF=\{M\}$.
We have seen one possible way to modify {the} 
utility function such that it satisfies 
$\{M\}$-VoI-fairness. Consider now the following class of utility functions 
$\Upsilon^*:=\{(D,S,N,M)\mapsto \mathbbm{1}(D=1)\left( w_1S+w_2N+w_3M\right) \mid (w_1,w_2,w_3)\in \mathbb{R}^3\}$.
Then,
    $\widetilde{\upsilon}(D,S,N,M):=
    \mathbbm{1}(D=1)(-\theta_S^{N'}\theta_{N'}^N\theta_N^U S+\theta_{N}^U N + (\theta_M^U+\cov(N,M)\theta_N^U/\text{var}(M))M)$
is the unique $\upsilon$-corresponding 
$\{M\}$-VoI-fair utility function in  $\Upsilon^*$, 
see~\Cref{app: analysolumedicalstaff} {(throughout the paper, we identify objects that are equal up to measure zero). }
This
$\widetilde{\upsilon}$ differs from the $\widetilde{\upsilon}$ in \Cref{ex: doctor VoI fair} only by the term 
$\frac{\cov(N,M)\theta_N^U}{\var(M)}M$. 
{  Since $\frac{\cov(N,M)\theta_N^U}{\text{var}(M)}>0$, this means that the $\upsilon$-corresponding $\{M\}$-VoI-fair utility in $\Upsilon^*$ weighs medical qualifications higher than the simple modification of the utility in \Cref{ex: doctor VoI fair}}.
In~\Cref{app: estsolmedstaff},
we %
obtain this $\upsilon$-corresponding
$\{M\}$-VoI-fair utility function based on data from an observed policy (using rejection sampling).

\subsubsection{College admission}\label{app: college admission}
\begin{table}[t]
\caption{
Synthetic experiment based on 
\Cref{app: college admission}.
We modify the utility function from 
\Cref{app: college admission}
by adding a term $-0.070S$, making the utility function $\{T\}$-VoI-fair.
We apply optimal policies under the original and modified utility function.
}
\centering
\begin{tabular}{@{}lcc@{}} \toprule
& \multicolumn{2}{c}{Utility functions} \\ \cmidrule(l){2-3}
 & Original & Modified \\ \midrule[0.6pt]
\begin{tabular}[c]{@{}l@{}}
Number of students graduating
\end{tabular}    & 25 670   & 25 573  \\ \midrule[0.1pt]
\begin{tabular}[c]{@{}l@{}}Number of students admitted\\ from minority group\end{tabular}   & 11 057   & 13 114\\ \midrule[0.1pt]
\begin{tabular}[c]{@{}l@{}}Number of students admitted and\\ graduating from minority group \end{tabular}  & 5 067  & 5 461\\
\bottomrule\end{tabular}
\label{table: college attainment}
\end{table}

We now consider the task of improving the utility function for college admission.
    Suppose we are choosing a subset of applicants for admission to a university's mathematics department. We have 
    three features for each applicant $i \in \{1, \ldots, n\}$ 
    on which we can base our policy, that is, 
    three features in $\rO^D$:
\ifsqueeze
  $S_i\sim \text{Bernoulli}\left(2/3\right)$: The protected attribute;  $T_i$: A score from a mathematics test taken as part of the admission process; and $R_i$: A score based on the applicant's resume. 
\else
    \begin{compactitem}
        \item[$\bullet$] $S_i\sim \text{Bernoulli}\left(2/3\right)$,  the protected attribute.  
        \item[$\bullet$] $T_i$: A score from a mathematics test taken as part of the admission process. 
        \item[$\bullet$] $R_i$: A score based on the applicant's resume. 
    \end{compactitem}
\fi
Based on the three features, we try to predict $Y_i$, representing whether applicant $i$ %
would
graduate if admitted.
    We assume that the minority group $S=0$ is the disadvantaged group. 
    For illustrative purposes, we assume that there are $n=100 \ 000$ 
    applicants and that we admit exactly half of them. %
    We assume that 
    $(S_1,T_1,R_1,Y_1), \ldots, 
    (S_n,T_n,R_n,Y_n)$
    are %
    identically distributed
    \ifsqueeze
    ,
        $Y\in \{0,1\}^n$, 
        $D=\pi(S,T,R)\in \{0,1\}^n$,
        and
        $U:=D^\top Y $
    \else
    and  
   $$
         Y\in \{0,1\}^n, \quad
        D=\pi(S,T,R)\in \{0,1\}^n, \quad
        U:=D^\top Y 
    $$
    \fi
    using notation $S := (S_1,..., S_n)$ and similar.
    Here, the policy is a function of the 
    entire feature vectors.
    We specify the data-generating model
    in \Cref{app: sim college example}.
Under the induced distribution, $S_i$ and $Y_i$ are not independent, that is, there is a difference 
    in graduation between the two groups. We do not interpret 
    the assignments in \Cref{app: sim college example} %
    as the causal mechanisms of the data-generating process 
    but only as a model for the observational distribution.

Assume for now that we choose $\rF =\{Y\}$ as the set of essential features. 
Then, $U$ is $\{Y\}$-VoI fair. 
The data-generating process 
in \Cref{app: sim college example} 
implies that
     for all $i, t, r$,
     \begin{align*}
        \mathbb{E}(Y_i \ | \  S_i=0, T_i=t,R_i=r)>
        \mathbb{E}(Y_i \ | \ S_i=1, T_i=t,R_i=r),
     \end{align*}
    that is, under an optimal policy based on $S$, $T$, and $R$, the university will, given $T_i=t$ and $R_i=r$, be more likely to admit a person from the minority group.
    We do not think that 
    this would be considered intuitively unfair.
    (Some may 
    interpret this 
    inequality 
    as the university 
    applying affirmative action, but this conclusion may be misleading since it is not a deliberate choice by the university but a consequence of optimally predicting graduation given the available data.) %
    However, if 
    resume scores $R_i$
    were unavailable, 
    $U$ is still $\{Y\}$-VoI fair but
    it turns out that  
    \ifsqueeze
    $ \mathbb{E}(Y_i \ | \ S_i=0, T_i=t)<\mathbb{E}(Y_i \ | \ S_i=1, T_i=t),$
    \else
    $ \mathbb{E}(Y_i \ | \ S_i=0, T_i=t)<\mathbb{E}(Y_i \ | \ S_i=1, T_i=t),$
    \fi
    that is, under an optimal policy based on $S$, and $T$, the university will, given $T_i=t$, be more likely to admit a person from the majority group.
 In other words, 
a $\{Y\}$-VoI-fair policy would use a higher test score threshold for admitting applicants from the minority group compared to the majority group if $\rO^D=\{S,T\}$. That the utility function creates this incentive may be considered intuitively unfair.
These comments
suggest that $\{T\}$, which avoids this incentive,
is a more appropriate choice for the set of essential features than $\{Y\}$.

Let us {now}, instead, consider $\rF =\{T\}$ as the set of essential features. Then, $U$ is not 
$\{T\}$-VoI fair (it is not $\{R,T\}$-VoI fair, either). We now
    find a $\{T\}$-VoI-fair utility function in the class $\Upsilon^*=\{D^\top\left(Y+wS\right) \mid  w\in \mathbb{R}\}$
by using simulated data from the data-generating process, see~\Cref{app: estimate w}.
When considering a specific realization of the data-generating process {and $\rO^D=\{S,T,R\}$}, the
{$U$-}corresponding 
$\{T\}$-VoI-fair utility function
in $\Upsilon^*$ 
    results in 97 fewer total graduates and 394 more graduates from the minority group (compared to the optimal policy under the original utility function). We provide all numbers in 
    \Cref{table: college attainment}; 
    \Cref{app: multiple realizations} 
contains results on different realizations.
There is a trade-off between avoiding intuitively unfair discrimination 
and a decrease in the total number of people graduating. In this example, VoI-fairness 
provides a principled approach for making the trade-off by requiring that $S$ does not have VoI relative to $\{T\}$. \citep{Nilforoshan2022} also advocate for adding a term $wS$ to the utility function, but, consider $\omega$ an 
``arbitrary constant that balances preferences for both student graduation and racial diversity.'' 
By letting test scores be the essential feature, we have a principled way to choose the offset $wS$ such that more persons from the minority group are admitted. By defining fairness in terms of the size of $\omega$, we avoid choosing policies outside of the Pareto frontier, that is, policies for which there exist alternatives that are superior both in terms of college attainment and diversity.

\subsection{Applying VoI-fairness to COMPAS data}
\label{ex: compas}

\begin{table}[h]
\caption{
Summary of results from \Cref{ex: compas}:
comparison of the optimal policies based on $\rO^D$
corresponding to the original 
and the modified 
$\{\text{charge degree}\}$-VoI-fair utility function.
The reported numbers are averages over the 20 runs (we
include the smallest/largest values obtained over the 20 runs in parentheses).
The two reported mean Brier scores, 0.205 and 0.215, are to be compared to 0.250, the mean Brier score of a constant policy which sets $D$ to be the empirical mean of $Y$ in the training fold.
}
\centering
\setlength{\tabcolsep}{10pt} %
\renewcommand{\arraystretch}{0.8} %
\begin{tabular}{@{}lcc@{}} 
\toprule
& \multicolumn{2}{c}{Utility functions} \\ 
\cmidrule(l){2-3}
 & Original & Modified \\ 
\midrule[0.6pt]
\begin{tabular}[c]{@{}l@{}}
$\lambda$
\end{tabular}    & 0    & 0.27 (0.19--0.35)  \\ 
\midrule[0.1pt]
\begin{tabular}[c]{@{}l@{}}
Mean pred. risk for \\ African-Americans
\end{tabular}   & 0.56 (0.55--0.58)   &  0.44 (0.40--0.48)\\ 
\midrule[0.1pt]
\begin{tabular}[c]{@{}l@{}} Mean pred. risk \\ for Caucasians \end{tabular}  & 0.43 (0.40--0.44)  & 0.43 (0.40--0.44)\\ 
\midrule[0.1pt]
\begin{tabular}[c]{@{}l@{}} Mean Brier score \end{tabular}  & 0.205 (0.197--0.213)  & 0.215 (0.202--0.228)\\ 
\midrule[0.1pt]
\begin{tabular}[c]{@{}l@{}} Mean Brier score \\ baseline \end{tabular}& \multicolumn{2}{c}{ 0.250 (0.250--0.251)} \\ 
\bottomrule
\end{tabular}
\label{table: compas}
\end{table}

COMPAS (Correctional Offender Management Profiling for Alternative Sanctions) is a 
proprietary risk-assessment tool that produces a decile risk score intended to
predict recidivism \citep{jackson2020setting}.
We use the publicly available dataset released by ProPublica \citep{larson2016compas}
and consider the following variables:
recidivism within 2 years, which we denote by $Y \in \{0,1\}$;
age; sex; 
race (for simplicity, we subset the dataset such that it only contains `African-American' and `Caucasian', which are the largest groups); count of prior offenses; charge degree (misdemeanor or felony); 
decile score (that is, a number ranging from 1 to 10, representing a defendant's likelihood of recidivism, as predicted by the COMPAS algorithm).

    We assume that $S = \{\text{race}\}$ is the protected attribute and that 
    $\rO^D = \{$age, sex, race, count of prior offenses, decile score, charge degree$\}$.%
\footnote{We can use the COMPAS decile score even though this score has been the subject of much controversy \citep{larson2016compas,corbett2016computer,jackson2020setting}: 
one of the main insights of our work is that a VoI-fair utility enables policies to use features that are seemingly `unfair' on their own,
see \Cref{ex: grade}.}
    The decision $D\in (0,1)$ is a predicted risk, which differs from the decile score. The 
 negative Brier score serves as a utility function, that is, 
        $\upsilon(D,Y)=-(D-Y)^2$.
    Finally,  we choose the set of essential features to be $\rF=\{\text{charge degree}\}$; this is based on the reasoning that it would be legitimate to use this feature even if it was substantially
    correlated with race.

We now, first (1), argue that the utility is not  $\{\text{charge degree}\}$-VoI-fair, second (2), propose a modified 
utility function that is $\{\text{charge degree}\}$-VoI-fair, and third (3), estimate the optimal policies under the original and modified utility and compare them.
We use XGBoost to estimate optimal policies, see \Cref{app: compas} for details.

(1) To test if the negative Brier score is $\{\text{charge degree}\}$-VoI fair, we split the data into two equally sized folds. We then estimate optimal policies based on $\{\text{charge degree}\}$ and $\{\text{charge degree},\text{race}\}$ using the first fold. 
Evaluating these on the second fold, we find that 
{  the Brier score is not a $\{\text{charge degree}\}$-VoI-fair utility: we obtain a higher utility using both charge degree and race compared to using only charge degree (given charge degree, being African-American increases the predicted risk of recidivism).}
(t-test: $p=0.003$).

    (2) To find a $\{\text{charge degree}\}$-VoI-fair utility, we consider utility functions of the form
    \begin{align*}
        &\Upsilon^*=\{(\rO^D,D,Y)\mapsto -(D-Y)^2\\
        &\qquad -\lambda \cdot D\cdot\mathbbm{1}(\text{African-American})\ \mid \lambda\in (0,\infty) \},
    \end{align*}
    where $\mathbbm{1}(\text{African-American})\in \{0,1\}$ is indicating if the subject is African-American. We want to choose $\lambda$ such that race does not have VoI relative to $\{\text{charge degree}\}$.
    We split the data into three folds (60\%, 20\%, 20\%). 
    We use the first fold to estimate optimal policies and the second fold to estimate the corresponding expected utilities (we use the third fold in (3)).
    This allows us to choose a $\lambda$ such that the resulting utility is $\{\text{charge degree}\}$-VoI-fair. {  There is existing work that modifies the training objective in order to push the predictions closer to complying with various fairness notions \citep{berk2017convex,bechavod2017penalizing}. These approaches raise the question of how to choose the strengths of the `unfairness penalty'. In our work, the strength of the regularization is determined by the VoI-fairness requirement.}

    (3) After having found a $\lambda$ to minimize the VoI of race relative to $\{\text{charge degree}\}$, we use 
    the first fold to train policies based on $\rO^D$ to maximize the original and modified utility functions
    and compare these using the third fold. 
    We repeat steps (2) and (3) 20 times using random splits and report the results in \Cref{table: compas} and \Cref{app: histograms}. (As a side note, in all 20 runs, the obtained utility is larger using $\{\text{charge degree, race}\}$ rather than just $\{\text{charge degree}\}$, see (1)). The code to reproduce these results 
is available on GitHub: \url{https://github.com/FrederikHJ/Unfair-Utilities}.

{  Adding an offset $\lambda$ to the utility  decreases how well the decision procedure performs on the original utility (mean Brier going from 0.205 to 0.215), while still performing better than the baseline constant estimator (mean Brier 0.250), see \Cref{table: compas}. This loss in accuracy is accompanied by a significant reduction in the predicted risk for African-Americans (mean predicted risk going from 0.56 to 0.44).}

\section{Summary and future work}\label{sec: future work}
We have proposed 
value of information fairness, 
a formal criterion that utility functions should satisfy 
and have provided corresponding graphical criteria. 
We have shown
how an unfair utility function 
could
be improved and have applied our framework to several synthetic and real-world examples.
The flow chart in 
\Cref{fig: decision procedure}
outlines a summary for handling questions of fairness based on our findings. VoI-fairness solves shortcomings of existing approaches to fairness as we show in 
\Cref{ex: parental status} and \Cref{sec: other workds}, and is tractable to apply in practice as we show in \Cref{sec: experiments}.
 \begin{figure}[t]
    \centering
\tikzset{noarrow/.style={red!60!black}}
\tikzset{yesarrow/.style={green!40!black,-{Latex[length=5mm, width=2mm]}]}}
   \begin{tikzpicture}[xscale = 1.1, yscale=1.2,thick,main/.style={draw,rectangle,minimum size=1cm}]
        \newcounter{tfn}
        \stepcounter{footnote}
        \setcounter{tfn}{\value{footnote}}
	\node[main,align=center] (1) at(0,2.5) {\textbf{1:} Pick a utility function $\upsilon$.\\Pick the features $\rF$ that \\ you consider essential.\\ Is the utility function $\rF$-VoI-fair?};
        \node[main,align=center] (2) at(5,2.5) {\textbf{2:} Do you consider the\\ optimal policy\\ ethically acceptable?\protect\footnotemark[\value{tfn}]\stepcounter{tfn}};
        \node[main,align=center] (3) at(0,0) {\textbf{3:} This utility function \\ should not be used.\\ Find a  $\rF$-VoI-fair utility \\ function,
     e.g., $\upsilon$-corresponding.\\
        Go to 2.};
        \node[main,align=center] (4) at(10,2.5) {\textbf{4:} Use this policy. \\ It is an $\rF$-VoI-fair policy.\protect\footnotemark[\value{tfn}]\stepcounter{tfn}};
        \node[main,align=center] (5) at(5,0) {\textbf{5:} Is it possible \\ to collect better data,  \\  e.g., more variables,
        for $\rO^D$?};
        \node[main,align=center] (6) at(10,0) {\textbf{6:} Try that. Go to 2.};
        \node[main,align=center] (7) at(5,-2.5) {\textbf{7:}Are you sure \\that the original\\ utility function is reasonable,\\
        and did you pick the \\ right essential features?\protect\footnotemark[\value{tfn}]\stepcounter{tfn}};
          \node[main,align=center] (8) at(10,-2.5) {\textbf{8:} Unclear what to do.\\ We are not sure whether \\there is a realizable policy\\ 
          that 
          can be agreed on to be\\
          strictly preferable over an\\ 
          $\rF$-VoI-fair policy.{\protect\footnotemark[\value{tfn}]}};
         \node[main,align=center] (9) at(0,-2.5) {\textbf{9:} Consider changing either\\ the original utility function or\\ the essential features.\\ Go to 1. };
        \draw[green!40!black,-{Latex[length=2.5mm, width=2.5mm]},line width=.7mm] (1) to node[above]{Yes} (2);
        \draw[green!40!black,-{Latex[length=2.5mm, width=2.5mm]},line width=.7mm] (2) to node[above]{Yes} (4);
        \draw[green!40!black,-{Latex[length=2.5mm, width=2.5mm]},line width=.7mm] (5) to node[above]{Yes} (6);
        \draw[green!40!black,-{Latex[length=2.5mm, width=2.5mm]},line width=.7mm] (7) to node[above]{Yes} (8);
        \draw[red!60!black,-{Latex[length=2.5mm, width=2.5mm]},line width=.7mm] (1) to node[left]{No} (3);
        \draw[red!60!black,-{Latex[length=2.5mm, width=2.5mm]},line width=.7mm] (2) to node[left]{No} (5);
        \draw[red!60!black,-{Latex[length=2.5mm, width=2.5mm]},line width=.7mm] (5) to node[left]{No} (7);
        \draw[red!60!black,-{Latex[length=2.5mm, width=2.5mm]},line width=.7mm] (7) to node[below]{No} (9);
\end{tikzpicture}
\caption{
Possible approach for handling decision problems with protected attributes. 
\label{fig: decision procedure}}
\end{figure}
\footnotetext[\value{footnote}]{
See \Cref{sec:ODnotsubsetM}. 
This paper does
not propose a formal criterion that policies must satisfy to be considered intuitively fair/ethically acceptable.
}
\stepcounter{footnote} \footnotetext[\value{footnote}]{Technically, one needs to make sure that the optimal policy does not use variables that do not have VoI, see \Cref{def: VoI-fair policy} and \Cref{app: VoI-fair policy}.}
\stepcounter{footnote} \footnotetext[\value{footnote}]{
In \Cref{ex: hair depends on M} the VoI-fair policy changes depending on which features we consider essential. 
In \Cref{ex: doctor VoI fair}, $U=\mathbbm{1}(D=1)\cdot (-M)$ is a $\{\text{Medical qualifications}\}$-VoI-fair utility function, but it would probably be considered intuitively unfair to have a preference for hiring unqualified doctors. In this case, one needs to reconsider the original utility function.}
\stepcounter{footnote} \footnotetext[\value{footnote}]{See \Cref{sec: VoI-fair policy is best}; see \Cref{ex: uncertainty} for why requiring additional fairness constraints may be undesirable.}

There are several open questions related to the framework we introduced. How can we infer a $\upsilon$-corresponding $\rF$-VoI-fair utility function
from observed data? (See \Cref{app: estsolmedstaff} for a simple example.) 
How can we collect data such that {VoI-fair} policies will be considered {ethically acceptable}? 
In some examples (see, e.g., \Cref{ex: hair}) the data were not sufficient to find an ethically acceptable policy; how can we guide the process of collecting further, helpful data?
Corresponding VoI-fair utility functions
can be defined using other metrics than \Cref{def: appropriate} (this  may be necessary if $D$ is continuous); what 
do different choices of metrics entail?
In practice, we may not be able to estimate optimal policies. To what extent does unfairness occur due to suboptimal policies rather than unfair utilities? {  We may want $S$ to be a vector of several protected attributes. How can we effectively find VoI-fair utilities in these cases? }

We believe that considering utility functions, rather than focusing exclusively on policies, 
adds an important 
piece to the puzzle of
algorithmic fairness 
and could help 
to reduce 
unwanted discrimination in 
real-world applications.

\section*{Authors' statements}
    
\subsection*{Acknowledgements}

We thank Niklas Pfister and  Jonathan Richens for helpful comments and discussions.

\subsection*{Funding information}
This research was supported by the Pioneer Centre for AI, DNRF grant number P1. During part of this project FHJ and JP were supported by a research grant (18968) from VILLUM FONDEN.

\subsection*{Author contributions}
All authors have accepted responsibility for the entire content of this 
manuscript and consented to its submission to the journal,
reviewed all the results and approved the final version of the manuscript.
FHJ conceived the initial idea and led the writing of the paper. 
SW and JP contributed with ideas, writing, supervision, and revising the paper.
\subsection*{Conflict of interest}
JP is a member of the Editorial Board of the Journal of Causal Inference but was not involved in the review process of this article.

\subsection*{Data availability statement}
We use the COMPAS dataset from the Julia Fairness package: 
\url{https://ashryaagr.github.io/Fairness.jl/dev/datasets/}.
Code to reproduce all experiments is available on GitHub: \url{https://github.com/FrederikHJ/Unfair-Utilities}.

\bibliography{references}

\newpage
\appendix

\section{Further details on VoI-fairness}
\subsection{Visual explanation of VoI-fairness}\label{app: visual}
\begin{figure}[h!]
    \centering
   \begin{tikzpicture}[yscale = 1.1, xscale=1,thick,main/.style={draw,rectangle,anchor=west},every node/.style={scale=1}]
        \node[align=left] (title) at (4,0) {Is $U$ 
        $\rF$-VoI-fair?};
	\node[main,fill=cyan] (D) at(.55,0) {$D$};
	\node[main,fill=yellow] (U) at(2,0) {$U$};
	\draw[->] (D) to (U);	
	\node[main] at(-2.3,1)  (b1) {Qualifications};
	\node[main,align=center] at(-2.3,0)  (b2) {Application\\[-.25em]material, $A$};
	\node[main] at(-2.3,-1)  (b3) {Parental leave, $S$};
	\node[main,align=center] at(-4.5,0) (h) {Unknown\\[-.25em]factors};
	\draw[->] (b1.south) to[*|] (b2);
	\draw[->] (h) to (b1.west);
	\draw[->] (h) to (b2.west);
	\draw[->] (h) to (b3.west);
	\draw[->,dashed,DarkRed] (b2) to (D);
	\node[main] at(1,1) (p1) {Quality of work, $Q$};
	\node[main] at(1,-1) (p2) {Work hours, $W$};
	\draw[->] (p1.south) to[*|] (U.north);
	\draw[->] (p2.north) to[*|] (U.south);
	
	\draw[->] (b1) to (p1);
	\draw[->] (b3) to (p2);

        \node [draw, rectangle, rounded corners, thick, darkgreen, fit=(p1)] {};
        \node [draw, rectangle, rounded corners, thick, DarkRed, fit=(b2)] {};

        \node [draw, circle, thick, DarkRed, align=left,text=black] at(4.5,-1) (od) {};
        \node [right=0.05cm of od] {$\rO^D$};
        \node [draw, circle, thick, darkgreen, align=left,text=black] at(4.5,-0.5) (F) {};
        \node [right=0.05cm of F] {$\rF$};

    \end{tikzpicture}
    \caption{This is a repetition of \Cref{fig: parental status}. Here, $U$ is not $\rF$-VoI fair. Figure~\ref{fig:777} explains why this is the case -- it serves as a visualization of the formal definitions. \label{fig:666}}
    
    \vspace{0.4cm}

   \begin{tikzpicture}[yscale = 1.1, xscale=1,thick,main/.style={draw,rectangle,anchor=west},every node/.style={scale=1}]
	\node[main,fill=cyan] (D) at(.55,0) {$D$};
	\node[main,fill=yellow] (U) at(2,0) {$U$};
	\draw[->] (D) to (U);	
	\node[main] at(-2.3,1)  (b1) {Qualifications};
	\node[main,align=center] at(-2.3,0)  (b2) {Application\\[-.25em]material, $A$};
	\node[main] at(-2.3,-1)  (b3) {Parental leave, $S$};
        \node[align=center] at (-1,-1.6) (comment) {Does $S$ have VoI?};
	\node[main,align=center] at(-4.5,0) (h) {Unknown\\[-.25em]factors};
	\node[main] at(1,1) (p1) {Quality of work, $Q$};
	\node[main] at(1,-1) (p2) {Work hours, $W$};
	\draw[->] (b1.south) to[*|] (b2);
	\draw[->] (h) to (b1.west);
	\draw[->] (h) to (b2.west);
	\draw[->] (h) to (b3.west);
	\draw[->,dashed,darkgreen] (p1) to (D);
 	\draw[->,dashed,transparent] (b3.north east) to [out=45,in=270] (D);
	\draw[->] (p1.south) to[*|] (U.north);
	\draw[->] (p2.north) to[*|] (U.south);
	
	\draw[->] (b1) to (p1);
	\draw[->] (b3) to (p2);

        \node [draw, rectangle, rounded corners,  thick, darkgreen, fit=(p1)] {};
        \node [draw, rectangle, rounded corners, thick, DarkRed, fit=(b2)] {};

        \node [draw, circle, thick, DarkRed, align=left,text=black] at(4.5,-1) (od) {};
        \node [right=0.05cm of od] {$\rO^D$};
        \node [draw, circle, thick, darkgreen, align=left,text=black] at(4.5,-0.5) (F) {};
        \node [right=0.05cm of F] {$\rF$};

    \end{tikzpicture}
    \caption{We imagine a hypothetical scenario where the decision inputs are $\rF=\{Q\}$ instead of $\rO^D$. In this hypothetical scenario, parental leave has value of information, that is, we could obtain a higher expected utility if parental leave was a decision input in addition to quality of work. Therefore, the utility is not $\{Q\}$-VoI fair. \label{fig:777}}

    \vspace{0.4cm}

       \begin{tikzpicture}[yscale = 1.1, xscale=1,thick,main/.style={draw,rectangle,anchor=west},every node/.style={scale=1}]
	\node[main,fill=cyan] (D) at(.55,0) {$D$};
	\node[main,fill=yellow] (U) at(2,0) {$\widetilde{U}$};
	\draw[->] (D) to (U);	
	\node[main] at(-2.3,1)  (b1) {Qualifications};
	\node[main,align=center] at(-2.3,0)  (b2) {Application\\[-.25em]material, $A$};
	\node[main] at(-2.3,-1)  (b3) {Parental leave, $S$};
	\node[main,align=center] at(-4.5,0) (h) {Unknown\\[-.25em]factors};
	\node[main] at(1,1) (p1) {Quality of work, $Q$};
	\node[main] at(1,-1) (p2) {Work hours, $W$};
	\draw[->] (b1.south) to[*|] (b2);
	\draw[->] (h) to (b1.west);
	\draw[->] (h) to (b2.west);
	\draw[->] (h) to (b3.west);
	\draw[->,dashed,darkgreen] (p1) to (D);
	\draw[->] (p1.south) to[*|] (U.north);
	\draw[->] (p2.north) to[*|] (U.south);
	
	\draw[->] (b1) to (p1);
	\draw[->] (b3) to (p2);
        \draw[->] (b3.north east) to (U);
        
        \node [draw, rectangle, rounded corners, thick, darkgreen, fit=(p1)] {};
        \node [draw, rectangle, rounded corners, thick, DarkRed, fit=(b2)] {};

        \node [draw, circle, thick, DarkRed, align=left,text=black] at(4.5,-1) (od) {};
        \node [right=0.05cm of od] {$\rO^D$};
        \node [draw, circle, thick, darkgreen, align=left,text=black] at(4.5,-0.5) (F) {};
        \node [right=0.05cm of F] {$\rF$};

    \end{tikzpicture}
    \caption{We modify the utility function such that we cannot obtain a higher expected utility by using parental leave as a decision input in addition to quality of work. Concretely, we 
    could
    use the utility function $\widetilde{U}=\widetilde{\upsilon}(D,Q,W,S)=D\cdot Q \cdot (W-\mathbb{E}(W\mid S))$. $\tilde{U}$ is $\{Q\}$-VoI fair.}

    \vspace{0.4cm}

   \begin{tikzpicture}[yscale = 1.1, xscale=1,thick,main/.style={draw,rectangle,anchor=west},every node/.style={scale=1}]
	\node[main,fill=cyan] (D) at(.55,0) {$D$};
	\node[main,fill=yellow] (U) at(2,0) {$\widetilde{U}$};
	\draw[->] (D) to (U);	
	\node[main] at(-2.3,1)  (b1) {Qualifications};
	\node[main,align=center] at(-2.3,0)  (b2) {Application\\[-.25em]material, $A$};
	\node[main] at(-2.3,-1)  (b3) {Parental leave, $S$};
	\node[main,align=center] at(-4.5,0) (h) {Unknown\\[-.25em]factors};
	\draw[->] (b1.south) to[*|] (b2);
	\draw[->] (h) to (b1.west);
	\draw[->] (h) to (b2.west);
	\draw[->] (h) to (b3.west);
	\draw[->,dashed,DarkRed] (b2) to (D);
	\node[main] at(1,1) (p1) {Quality of work, $Q$};
	\node[main] at(1,-1) (p2) {Work hours, $W$};
	\draw[->] (p1.south) to[*|] (U.north);
	\draw[->] (p2.north) to[*|] (U.south);
	
	\draw[->] (b1) to (p1);
	\draw[->] (b3) to (p2);
        \draw[->] (b3.north east) to (U);

        \node [draw, rectangle, rounded corners, thick, darkgreen, fit=(p1)] {};
        \node [draw, rectangle, rounded corners, thick, DarkRed, fit=(b2)] {};

        \node [draw, circle, thick, DarkRed, align=left,text=black] at(4.5,-1) (od) {};
        \node [right=0.05cm of od] {$\rO^D$};
        \node [draw, circle, thick, darkgreen, align=left,text=black] at(4.5,-0.5) (F) {};
        \node [right=0.05cm of F] {$\rF$};

    \end{tikzpicture}
    \caption{A $\{Q\}$-VoI fair policy, see \Cref{def: VoI-fair policy}, is an optimal policy under the modified utility function, using the usable decision inputs $\rO^D$, that is, application material. \label{fig:999}}
\end{figure}

\subsection{On the definition of VoI-fair policies}\label{app: VoI-fair policy}
\begin{figure}[t]
    \centering
    \begin{tikzpicture}[node distance={20mm},thick,main/.style={draw,rectangle},scale=0.5]
    \node[fill=none,style=main](S) at (0,-2) {$S$};
    \node[fill=none,style=main](E) at (-0.5,0) {Abilities};
    \node[fill=cyan,style=main](D) at (2,-1) {$D$};
    \node[fill=yellow,style=main](U) at (4,-1) {$U$};
    \draw[->,dashed,DarkRed] (S) to (D);
    \draw[->,looseness=1] (E) to (U);
    \draw[->] (D) to (U);
        \end{tikzpicture}
    \caption{
    This graph visualizes the example in \Cref{app: VoI-fair policy}. Here, we can construct an optimal policy depending on $S$, but this seems intuitively unfair. Since there also exist optimal policies in $\Pi^\emptyset$ that do not use $S$, we require %
    that an $\{\text{Abilities}\}$-VoI-fair policy is in $\Pi^{\emptyset}$, %
    see \Cref{def: VoI-fair policy}.}%
    \label{fig: VoI-fair policy example}
\end{figure}
We now provide an example to illustrate why we require that all variables $X\in \rH\subseteq \rO^D$ have VoI relative to $\rH \backslash \{X\}$ in \Cref{def: VoI-fair policy}. 
We again consider a setting where $A$ indicates abilities and $D \in \{0,1\}$ is college admission.

\begin{align*}
    &S:{=\varepsilon_S}\sim \text{Unif}(\{0,1\})\\
    &A:=\varepsilon_{A} \sim \calN(0,1)\\
    &U:=\mathbbm{1}(D=1)\cdot A
\end{align*}
Assume that $\rO^D=\{S\}$, see \Cref{fig: VoI-fair policy example}. $U$ is \{\text{Abilities}\}-VoI-fair. 
\begin{align*}
    \pi(S)=1(S=1)
\end{align*}
is an optimal policy in $\Pi^{\rO^D}$, 
but in this example, it seems intuitively unfair to base the decisions on $S$ since $S$ does not have VoI relative to $\emptyset$. Therefore, in this example, if $\pi$ is an $\{\text{Abilities}\}$-VoI-fair policy, $\pi\in \Pi^{\emptyset}$.

If there is an optimal policy in $\Pi^{\rO^D}$, then it is always possible to find a subset $\rH \subseteq \rO^D$ and a policy $\pi\in \Pi^{\rH}$ such that every variable $X\in \rH$ have VoI relative to $\rH \backslash \{X\}$ and $\pi$ is optimal in $\Pi^{\rO^D}$. This follows immediately from the definition of VoI. {  Even when $S\notin \rO^D$, the requirement that all  variables in $\rH$ have VoI may be relevant to prevent arbitrary discrimination based on features correlated with $S$.
}

\subsection{
A VoI-fair policy may be a realizable policy closest to the unrealizable oracle policy
}\label{sec: VoI-fair policy is best}
In \Cref{ex: doctor VoI fair}, the oracle policy~\eqref{eq:orpol2} under the modified utility $\widetilde{U}$ is intuitively fair but not realizable.
Now, consider a 
general
situation where the 
utility function has the form $U=\mathbbm{1}(D=1)\widehat{U}$
for some function 
$\widehat{U}$
and $U$ is $\rF$-VoI-fair for 
essential features $\rF\subseteq \rX\backslash (\DE^D \cup \{S\})$.
We propose to measure the distance of a
policy
to the oracle policy as
\ifsqueeze
    $\delta :=\mathbbm{1}(D=1)\mathbbm{1}(\widehat{U}<0)|\widehat{U}|+\mathbbm{1}(D=0)\mathbbm{1}(\widehat{U}>0)\widehat{U}$.
\else
\begin{align}%
    \delta :=\mathbbm{1}(D=1)\mathbbm{1}(\widehat{U}<0)|\widehat{U}|+\mathbbm{1}(D=0)\mathbbm{1}(\widehat{U}>0)\widehat{U}.
\end{align}
\fi
This quantity may be called \textit{undesert}. {  In the context of hiring, it measures how big `mistakes' are made, that is, hiring people who perform poorly or not hiring people who perform well, weighted by how far the utility is from the indifference threshold 0.}
Since $\delta=-\mathbbm{1}(D=1)\widehat{U}+\mathbbm{1}(\widehat{U}>0)\widehat{U}$, 
an optimal 
policy minimizes the expected amount of undesert.
Thus, no realizable policy is better than a VoI-fair policy from the perspective of minimizing undesert.
It is possible that
a VoI-fair policy yields 
\begin{align}\label{eq: 1}
    \mathbb{E}\left(\delta \ | \ S=1\right) \neq 
    \mathbb{E}\left(\delta \ | \ S=0\right),
\end{align}
for example, %
which may seem intuitively unfair: Members of one group receive more undeserved treatment in expectation, but as long as $S\in \rO^D$
both sides of the inequality in \Cref{eq: 1} 
are minimized by a VoI-fair policy.

\section{Further details on examples}
\subsection{Overview of examples}\label{app: table examples}
\Cref{table: examples} provides an overview of all examples provided in this paper. 
\begin{table}[ht]
\centering
\caption{Overview of the examples} \label{table: examples}
\begin{tabular}{@{}ll@{}} \toprule
\Cref{ex: parental status}  & \begin{tabular}{@{}l@{}}
     Introductory example.
\end{tabular}     \\ \midrule
\Cref{ex: grade} &  \begin{tabular}{@{}l@{}}
     Illustrates that a VoI-fair policy may need to be a function of $S$. 
\end{tabular}  \\ \midrule
 & \begin{tabular}{@{}l@{}}
 $\rO^D$ may be such that no realizable policy, $\pi\in \Pi^{\rO^D}$, can
be considered intuitively fair.     \\ We provi{d}e four examples:
\end{tabular} \\ \cmidrule(l){2-2}
\Cref{ex: hair} &  \begin{tabular}{@{}l@{}}
     Illustrates unfairness due to unavailability of relevant data.
\end{tabular}  \\ \cmidrule(l){2-2}
\Cref{ex: uncertainty} &  \begin{tabular}{@{}l@{}}
     Illustrates unfairness due to unequal data quality.\\
     We also discuss the implications of enforcing\\ demographic parity, equalized odds, and counterfactual fairness.
\end{tabular}  \\ \cmidrule(l){2-2}
\Cref{ex: unfair no S} &  \begin{tabular}{@{}l@{}}
     Illustrates unfairness due to unavailability of $S$.\\
\end{tabular}  \\ \cmidrule(l){2-2}
\Cref{ex: entangled discrimination} &  \begin{tabular}{@{}l@{}}
     Illustrates unfairness due to entangled unwanted discrimination in data.\\
\end{tabular}  \\ \midrule
\Cref{ex: CF} &  \begin{tabular}{@{}l@{}}
     Shows how VoI-fairness may solve shortcomings of
counterfactual fairness.
\end{tabular}  \\ \midrule
\Cref{ex: EO} &  \begin{tabular}{@{}l@{}}
   Shows how VoI-fairness may solve shortcomings of
    equalized odds.
\end{tabular}  \\ \midrule
\Cref{ex: doctor VoI fair} &  \begin{tabular}{@{}l@{}}
     Shows how a human bias may make a utility function unfair and\\
     discusses a modification that makes the utility function VoI-fair. 
\end{tabular}  \\ \midrule
\Cref{ex: doctor VoI fair continued} &  \begin{tabular}{@{}l@{}}
     Provides the $\upsilon$-corresponding VoI-fair utility function for \Cref{ex: doctor VoI fair}.\\
\end{tabular}  \\ \midrule
\Cref{app: college admission}
&  \begin{tabular}{@{}l@{}}
     Illustrates a possible improvement of the utility function for college admission \\
     by using test scores as the essential feature.
\end{tabular}  \\ \midrule
\Cref{ex: compas} &  \begin{tabular}{@{}l@{}}
    We apply VoI-fairness on the COMPAS data, suggesting a modified utility \\ that 
disincentivizes large predicted risks for African Americans.
\end{tabular}  \\ \midrule
\Cref{ex: hair depends on M} &  \begin{tabular}{@{}l@{}}
     Shows that the $\upsilon$-corresponding VoI-fair utility function depends on the choice of $\rF$. \\
     If the VoI-fair policy seems intuitively unfair, it may be that you should reconsider $\rF$.
\end{tabular}  \\
\bottomrule \end{tabular}
\end{table}

\subsection{Further examples for Section \ref{sec:ODnotsubsetM} 
}\label{app: unfair}
\begin{example}[\textbf{Unfairness due to unequal data quality%
}]\label{ex: uncertainty}
        Consider the structural assignments
        \begin{align*}
	&S:=\varepsilon_S\sim\text{Ber}(0.99) \\
        &A:=\varepsilon_A\sim \mathcal{N}(0,1)\\
	&G:=\mathbbm{1}(S=0)(A+\sqrt{2}\varepsilon_{G_0}) + \mathbbm{1}(S=1)(A+\varepsilon_{G_1}) \\ 
        &U:=\mathbbm{1}(D=1)\cdot (A-1),
	\end{align*}
with $\varepsilon_{G_0},\varepsilon_{G_1}\sim \calN(0,1)$
        and assume that the usable decision inputs are $\rO^D=\{G,S\}$. The situation is similar to \Cref{ex: grade}, but now, the grade 
        $G$
        is a noisy function of me abilities, {with larger noise variance} for group $S=0$ than for group $S=1$. 
        $U$ is $\{A\}$-VoI-fair and 
        \begin{align} \label{eq: VoI-fair policy} %
            \pi(G,S)=\mathbbm{1}(S=1)\mathbbm{1}(G\geq
		2)+\mathbbm{1}(S=0)\mathbbm{1}(G\geq 3)
        \end{align}
is an $\{A\}$-VoI-fair policy since it is optimal in $\Pi^{\rO^D}$ (and no policy using only either $G$ or $S$ is optimal). 
            The same arguments apply if the groups have similar sizes. This policy seems intuitively unfair since $S\indep A$ but $\mathbb{P}(G \geq 2 | S=1)>\mathbb{P}(G\geq 3 | S=0)$.
            It is not clear if there is a better realizable policy in $\Pi^{\rO^D}$. 
            In \Cref{app: consequences}, we consider what would happen if we enforce demographic parity, equalized odds, or counterfactual fairness in this example. In summary, they all suggest different solutions than~\eqref{eq: VoI-fair policy} and can be considered intuitively unfair, too.
        \end{example}
\begin{example}{\textbf{(Unfairness due to unavailability of $S$).}}\label{ex: unfair no S}
Consider \Cref{ex: grade} but with $\rO^D=\{\text{Grade}\}$.
Then, the policy  $\pi(G)=\mathbbm{1}(G\geq 0)$ 
is $\{\text{Abilities}\}$-VoI-fair but seems unacceptable.
This policy does not manage to adjust for the unwanted discrimination $\alpha S$ in grade. In this case, it seems that no realizable policy in $\Pi^{\rO^D}$ is ethically acceptable, and 
we suggest that we need to collect better data (see \Cref{fig: decision procedure}). 

This example also 
highlights a known fundamental problem with fairness through unawareness: 
If the policy is not allowed to be a function of $S$, this may reduce our ability to choose an intuitively fair policy
because we cannot adjust for biases in the data
(a similar point is made in \citep{Kusner2017}).
\end{example}
\begin{example}[\textbf{Unfairness due to entangled unwanted discrimination in data}]\label{ex: entangled discrimination}
    Consider again \Cref{ex: grade} with $\rO^D=\{\text{Grade},S\}$
    and
    \begin{align*}
        G:=\mathbbm{1}(S=1)\cdot A&+\mathbbm{1}(S=-1)\mathbbm{1}(A>0)\cdot (-A)+\mathbbm{1}(S=-1)\mathbbm{1}(A\leq 0)\cdot A,
    \end{align*}
    corresponding to a situation where people from group $S=-1$ with $A>0$ are experiencing unwanted discrimination. If we, for example, observe $(S=-1, G=-1)$, the posterior will be $\mathbb{P}(A=1 \ | \ S=-1, G=-1)=\mathbb{P}(A=-1 \ | \ S=-1, G=-1)=0.5$.
    Any optimal policy in $\Pi^{\rO^D}$ is $\{\text{Abilities}\}$-VoI-fair. Thus, some policies are $\{\text{Abilities}\}$-VoI-fair even though they do not admit anyone from group $S=-1$. In this example, we consider it unclear what the intuitively fairest policy in $\Pi^{\rO^D}$ is.
\end{example}

\subsection{Further details on \Cref{ex: hair}}\label{app: hair depends on M}
\begin{example}{\textbf{(The essential features may affect which utility functions are $\upsilon$-corresponding $\rF$-VoI-fair).}}\label{ex: hair depends on M}
Consider the setting from \Cref{ex: hair}. If we 
consider physical strength
essential for the task, $\rF=\{P\}$, then a $\{P\}$-VoI-fair policy uses hair length to infer physical strength. 
If physical strength is not considered an essential feature,
the $\upsilon$-corresponding $\emptyset$-VoI-fair utility function in
$\Upsilon^*:=\{(D,S,H,P)\mapsto  \mathbbm{1}(D=1)\left(w_1S+w_2H+w_3P\right) \mid (w_1,w_2,w_3)\in \mathbb{R}^3\}$
is %
$\widetilde{U}:=\widetilde{\upsilon}(D,S,P) := 
\mathbbm{1}(D=1)
(P-\theta_S^P S)$.
All $\emptyset$-VoI-fair policies under $\widetilde{\upsilon}$ are 
policies in~$\Pi^{\emptyset}$. To summarize, if we do not consider physical strength an essential feature, VoI-fairness recommends a constant policy. Therefore, if one thinks that the VoI-fair policy in \Cref{ex: hair} is intuitively unacceptable, that may be because one does not 
consider physical strength an essential feature.
\end{example}

\subsection{Further details on Example \ref{ex: uncertainty}}\label{app: consequences}
\subsubsection{Consequences of fairness constraints}
        \textbf{Equalized odds and counterfactual fairness:}
        The policy in (\ref{eq: VoI-fair policy}) does not satisfy some existing definitions of fairness. E.g., it does  not satisfy 
        \begin{align}\label{eq: equal odds}
            S \indep D \mid A,
        \end{align}
        which is sometimes called counterfactual equalized odds \citep{Coston2020}. It does not 
        satisfy counterfactual fairness \citep{Kusner2017},
        \begin{align}\label{eq: counterfactual fairness}
           \mathbb{P}(D_{S:=0}=1\mid G,S)=\mathbb{P}(D_{S:=1}=1\mid G,S)
        \end{align}
        (with $D_{S:=0}$ and
        $D_{S:=1}$ being the counterfactuals of $D$ had $S$ been $0$ or $1$, respectively),
        either.
        If a realizable policy $\pi\in \Pi^{\rO^D}$ is to satisfy counterfactual fairness, then $\pi\in \Pi^{\emptyset}$, that is, the policy 
        is purely random
        (see \Cref{app: CF} for an argument in an analogous case).
        If we assume a threshold policy, $\pi(G,S)=\mathbbm{1}(S=1,G\geq c_1)+\mathbbm{1}(S=0,G\geq c_0)$ for $c_1,c_0\in \mathbb{R}\cup \{-\infty,\infty\}$,
        based on $\rO^D=\{S,G\}$, then equalized odds can also only be satisfied if $\pi \in \Pi^{\emptyset}$ (see \Cref{app: consequences EO}).
        Since $\mathbb{E}(A-1)<0$, the optimal policy under either equalized odds or counterfactual fairness would be $\pi(G,S)=0$, corresponding to not admitting any students. 
  While this might be a good meta-incentive to collect better data, it does not seem like a serious answer to the question of what should be done given the data available. It is possible to satisfy equalized odds using a threshold policy by adding extra independent noise $\varepsilon^*\sim \mathcal{N}(0,1)$ to the grade measurement for group $S=1$ such that $\text{var}(\varepsilon_{G_1}+\varepsilon^*)=\text{var}(\sqrt{2}\varepsilon_{G_0})$. This increases the amount of undesert (see \Cref{sec: VoI-fair policy is best}) in group $S=1$, $\mathbb{E}(\delta \ | \ S=1)$,  without affecting group $S=0$ \citep{Mittelstandt2023}.
        
        \textbf{Demographic parity:} It may seem that since $A \indep S$, an intuitively fair policy must satisfy demographic parity, that is, $S\indep D$. 
The only way to satisfy demographic parity would be to either admit
 people from group $S=0$ that are unqualified (in expectation) or reject people from group $S=1$ that are qualified (in expectation). 
        This is especially counterintuitive if $\text{var}(\varepsilon_{G_0})\gg\text{var}(\varepsilon_{G_1})$. 
        We would either have to admit candidates from group $S=0$ who are unlikely to be qualified or reject candidates from group $S=1$ who are likely qualified.
 Demographic parity would be an attractive property to satisfy in this example (indeed, an oracle policy would satisfy demographic parity), but we do not think that this is necessarily 
something that one should force upon the algorithm by doing constrained optimization.
        When we have to choose a realizable policy $\pi\in \Pi^{\rO^D}$, it may be
too late to satisfy demographic parity. 
        We think that future work should investigate how to collect data such that optimal policies under VoI-fair 
        utility functions result in intuitively fair policies. In this example, it seems that we should try to reduce $\text{var}(\varepsilon_{G_0})$.

\subsubsection{Further details on counterfactual equalized odds}\label{app: consequences EO}

Assume that $\pi\in \Pi^{\{G,S\}}$ satisfies counterfactual equalized odds,
    \begin{align*}
        \pi(G,S) \indep S \mid A
    \end{align*}
    Assume that $\pi$ is deterministic and that
there are constants $c_0,c_1\in \mathbb{R}\cup \{-\infty,\infty\}$ such that 
    \begin{align*}
        \pi(g,1)= \begin{cases}
        1 & g\geq c_1\\
        0 & g< c_1\
        \end{cases}\\
        \pi(g,0)=\begin{cases}
        1 & g\geq c_0\\
        0 & g< c_0\
        \end{cases}
    \end{align*}
    First, assume that $c_1,c_0\in \mathbb{R}$. Counterfactual equalized odds implies that
    \begin{align*}
    \exists a \in \mathbb{R} \,:\, a > \max(c_1, c_0) \; \text{ and } \; 
        \mathbb{P}(\varepsilon_{G_1}\in [c_1-a,\infty))=\mathbb{P}( \sqrt{2}\varepsilon_{G_0}\in [c_0-a,\infty)).
    \end{align*}
    Since $\text{var}(\sqrt{2}\varepsilon_{G_0})>\text{var}(\varepsilon_{G_1})$, 
    this implies that $c_0<c_1$. Likewise, 
    \begin{align*}
    \exists a \in \mathbb{R} \,:\, a < \min(c_1,c_0) \; \text{ and } \; 
        \mathbb{P}(\varepsilon_{G_1}\in [c_1-a,\infty))=\mathbb{P}(\sqrt{2}\varepsilon_{G_0}\in [c_0-a,\infty)),
    \end{align*}
    but now this implies $c_0>c_1$. So either $c_0\in \{-\infty,\infty\}$ or $c_1\in \{-\infty,\infty\}$.  Counterfactual equalized odds implies that $c_0=\infty$ if and only if $c_1=\infty$,  and similarly, $c_0=-\infty$ if and only if $c_1=-\infty$. So $\pi\in \Pi^{\emptyset}$.

\subsection{Further details on Example \ref{ex: CF}}\label{app: CF}
    We consider the setup from \Cref{ex: CF} and will show that counterfactual fairness implies choosing a policy in $\Pi^{\emptyset}$. 
    Following Proposition F.8 in \citep{Nilforoshan2022}, 
    it suffices to argue
   that 
    \begin{enumerate}
        \item For all $s\in \{-1,1\}$ and events $V$ satisfying $\mathbb{P}((G,S)\in V \mid S=s)>0$, we have that 
        \begin{align*}
            \mathbb{P}((G,S)_{S:=s}\in V \vee S=s \mid G=g,S=s')>0.
        \end{align*}
        for $\mathbb{P}_{(G,S)}$-almost all $g\in \mathbb{R}$ and
        $s'\in \{-1,1\}$.
     \item For all $s\in \{-1,1\}$ and $\epsilon>0$, there exists $\delta>0$ such that for any event $V$ satisfying $\mathbb{P}((G,S)\in V\mid S=s)<\delta$, we have that
        \begin{align*}
            \mathbb{P}((G,S)_{S:=s}\in V,S\neq s \mid G=g,S=s')<\epsilon.
        \end{align*}
        for $\mathbb{P}_{(G,S)}$-almost all $g\in \mathbb{R}$ and
        $s'\in \{-1,1\}$.
   \end{enumerate}

    Consider statement 1. Let $s\in \{-1,1\}$ and event $V$ be given such that $\mathbb{P}((G,S)\in V \mid S=s)>0$. Clearly,  $\mathbb{P}((G,S)_{S:=s}\in V \vee S=s \mid G=g,S=s)>0$ for almost all $g\in \mathbb{R}$, so it suffices to argue that $\mathbb{P}((G,S)_{S:=s}\in V \vee S=s \mid G=g,S \neq s)>0$ for almost all $g\in \mathbb{R}$. For almost all $g\in \mathbb{R}$, we have that 
    \begin{align*}
        \mathbb{P}((G,S)_{S:=s}\in V \vee S=s \mid G=g,S \neq s)&=\mathbb{P}((G,S)_{S:=s}\in V \mid G=g,S \neq s)\\
        &=\mathbb{P}((G_{S:=s},s)\in V \mid G=g,S \neq s)\\
        &=\mathbb{P}(G_{S:=s}\in V_s \mid G=g,S \neq s),
    \end{align*}
    where $V_s=\{g\in \mathbb{R}  \mid (g,s)\in V\}$ is Borel measurable, see, for example, \citep{Lauritzen2019}. Likewise, we have that $\mathbb{P}((G,S)\in V \mid S=s)=\mathbb{P}(G\in V_s \mid S=s)$. The statement now follows from the fact that for almost all $g\in \mathbb{R}$, $G_{S:=s} \mid (G=g,S\neq s)$ and $G\mid S=s$ are both normally distributed 
    (and therefore have the same null sets).

    Consider now statement 2. Let $s\in \{-1,1\}$ and $\epsilon>0$ be given. $\mathbb{P}((G,S)_{S:=s}\in V,S\neq s \mid G=g,S=s)=0$ for almost all $g\in \mathbb{R}$, 
    so it suffices
    to find a $\delta>0$ such that $\mathbb{P}((G,S)_{S:=s}\in V \mid G=g,S\neq s)<\epsilon$ for almost all $g\in \mathbb{R}$ whenever $\mathbb{P}((G,S)\in V\mid S=s)<\delta$. 
    An argument analogous to the one for statement 1 implies
    that $\mathbb{P}((G,S)_{S:=s}\in V \mid G=g,S\neq s)=\mathbb{P}(G_{S:=s}\in V_s \mid G=g,S\neq s)$ and $\mathbb{P}((G,S)\in V\mid S=s)=\mathbb{P}(G\in V_s\mid S=s)$. The statement now follows by applying Lemma 4.2.1 in \citep{Cohn2013}  {  (which characterizes absolute continuity of measures in terms of the $\epsilon-\delta$ condition we are aiming to show)}, since for almost all $g\in \mathbb{R}$, $G_{S:=s} \mid (G=g,S\neq s)$ and $G\mid S=s$ are both normally distributed and are therefore equivalent measures. 

\subsection{Path-specific counterfactual fairness in Example \ref{ex: doctor VoI fair}}\label{app: path-specific}
The $\{M\}$-VoI fair policy $\widetilde{\pi}=\mathbbm{1}\left(\theta_{M'}^M\theta_M^U M'+\theta_{N'}^N\theta_N^U N'-\theta_S^{N'}\theta_{N'}^N\theta_N^U S \geq 0\right)$ satisfies path-specific counterfactual fairness \citep{Nilforoshan2022, Wu2019} 
when considering
$\mathcal{M}=\{S\to D,S\to N' \to D\}$ as unfair paths.
In our notation,
we say that a policy satisfies path-specific counterfactual fairness if 
\begin{align}
\mathbb{P}(D_{S:=-1;\mathcal{M}}=1\mid \rO^D)=\mathbb{P}(D_{S:=1;\mathcal{M}}=1\mid \rO^D),
\end{align} 
where the counterfactuals are propagated only through unfair paths $\mathcal{M}$, see \citep{Pearl2001}
for a precise definition.
That $\widetilde{\pi}$ satisfies path-specific counterfactual fairness
is implied by
the counterfactual interviewers' evaluation 
(had $S$ been $-1$)
having point mass: $\mathbb{P}(N'_{S:=-1}=n-2\theta_S^{N'}|N'=n,S=1)=1$, 
which is arguably unrealistic. 

Assume now that we have
the structural assignment $$N'=\mathbbm{1}(S=-1)\varepsilon_{1}+\mathbbm{1}(S=1)\varepsilon_{2},$$ with $\begin{pmatrix}
    \varepsilon_1\\
    \varepsilon_2
\end{pmatrix}\sim \mathcal{N}\left(\begin{pmatrix}0\\0 \end{pmatrix}, \begin{pmatrix}
    1 & \sigma\\
    \sigma & 1 
\end{pmatrix}\right)$, $\sigma\in (-1,1)$. Then the original utility $U$ is 
$\{M\}$-VoI fair, and $\pi(M',N',S)=\mathbbm{1}\left(\theta_{M'}^M\theta_M^U M'+\theta_{N'}^N\theta_N^U N' \geq 0\right)$ is the corresponding VoI-fair policy, but $\pi$ does not satisfy path-specific counterfactual fairness. 
Furthermore, the results in \citep{Nilforoshan2022}
suggest (formally, this is 
proven only for discrete random variables) that
any realizable policy $\pi\in \Pi^{\rO^D}$ that satisfies path-specific counterfactual fairness with this new assignment of $N'$, would need to be a policy in $\Pi^{\{M'\}}$.
Intuitively, the constraint imposed by path-specific counterfactual fairness seems too strong since there is now no systemic bias in the interviewers' evaluations.

Similarly, if $$N'=\theta_S^{N'}S+\mathbbm{1}(S=-1)\varepsilon_{1}+\mathbbm{1}(S=1)\varepsilon_{2},$$
$\widetilde{\upsilon}(D,M,N,S)$, as given in \Cref{ex: doctor VoI fair}, would still be $\{M\}$-VoI-fair (and $\widetilde{\pi}$ an $\{M\}$-VoI-fair policy), but, as above, the results in \citep{Nilforoshan2022} suggest that only a policy in $\Pi^{\{M'\}}$ can satisfy path-specific counterfactual fairness. 
Again, 
path-specific counterfactual fairness
seems too restrictive;
we consider it preferable to correct for unwanted bias (as suggested by the framework
of VoI-fairness),
yielding an intuitively fair policy 
under which the expected value of 
the
utility $\widetilde{U}$ is strictly larger.

Instead of 
path-specific counterfactual fairness one 
can also 
consider path-specific interventional fairness \citep{Nabi2018,Chiappa2019,Kilbertus2017}. We say that a policy satisfies path-specific interventional fairness with unfair paths $\mathcal{M}$ if
\begin{align}
\mathbb{P}(D_{S:=-1;\mathcal{M}}=1)=\mathbb{P}(D_{S:=1;\mathcal{M}}=1).
\end{align} 
This
is satisfied by the $\{M\}$-VoI fair policy $\widetilde{\pi}$ with unfair paths $\mathcal{M}=\{S\to D,S\to N' \to D\}$. In general, it is not the case that an $\rF$-VoI-fair policy satisfies path-specific interventional fairness (unfair paths being all directed paths from $S$ to $D$ that do not pass through nodes $\rF$), see \Cref{ex: uncertainty} for a counterexample.\footnote{  
Strictly speaking,
the present example is also a counterexample since we consider $\{M\}$ rather than $\{M'\}$ to be the set of essential features. Yet, it seems that not including $S\to M'\to D$ in the set of unfair paths is the most faithful translation into path-specific fairness. }
Conversely,
a policy can satisfy path-specific interventional fairness without satisfying VoI-fairness. Consider, for example, that $S$ is not a parent of $N'$ but rather connected via a confounder, see \Cref{fig: confounder}. Assume further that the distribution over $(S,N',M')$ is unchanged. Then the optimal policy $\pi(M',N',S)=\mathbbm{1}\left(\theta_{M'}^M\theta_M^U M'+\theta_{N'}^N\theta_N^U N' \geq 0\right)$ under the original utility $U$ satisfies path-specific interventional fairness with $\mathcal{M}=\{S\to D\}$, but $\pi$ is not $\{M\}$-VoI-fair, as discussed in \Cref{ex: doctor VoI fair}. Here, path-specific interventional fairness is satisfied even though the policy is intuitively unfair.

Another problem with path-specific fairness 
may arise when
 some causal effects via a path may be considered fair while others will be considered unfair. Consider the modification depicted in \Cref{fig: noMprime}, where we assume that $M'$ is unobserved. Generally, the direct path $S\to D$ must be considered unfair to disallow direct unwanted discrimination. The path $S\to N'\to D$ is also considered unfair because of the unwanted bias in interviewer evaluations. But $\mathcal{M}=\{S\to D,S\to N'\to D\}$ also precludes that $S$ is used as a predictor of $M$. An $\{M\}$-VoI fair policy would use $S$ to adjust for the human bias an $N$ and to predict $M$ (which we find is the intuitively right thing to do here).    

In general, we can apply VoI-fairness without discovering the causal structure, see \Cref{sec: experiments}. Finally,  VoI-fairness 
avoids difficult 
philosophical issues involved in making sense of interventions on protected attributes \citep{sen2016, Hu2020,Kasirzadeh2021, mosse2025modeling}. 
As such, we regard VoI-fairness 
as a useful alternative to both 
path-specific interventional and counterfactual fairness.
    \begin{figure}[t]
		\centering
		\begin{tikzpicture}[node distance={20mm},thick,main/.style={draw,rectangle},scale=0.5]
		\node[fill=none,style=main](S) at (0,-2) {$S$};
		\node[fill=none,style=main](Mp) at (3,-.3) {$M'$};
            \node[fill=none,style=main](W) at (0,-3.7) {$W$};
		\node[fill=none,style=main](Np) at (3,-3.7) {$N'$};
		\node[fill=none,style=main](M) at (6,-.3) {$M$};
		\node[fill=none,style=main](N) at (6,-3.7) {$N$};
		\node[fill=cyan,style=main](D) at (4.5,-2) {$D$};
		\node[fill=yellow,style=main](U) at (9,-2) {$U$};
		\draw[->] (S) to (Mp);
		\draw[->] (W) to (S);
            \draw[->] (W) to (Np);
		\draw[->,dashed,DarkRed] (Mp) to (D);
		\draw[->] (M) to (U);
		\draw[->,dashed,DarkRed] (Np) to (D);
		\draw[->] (N) to (U);
		\draw[->] (D) to (U);
		\draw[->,dashed,DarkRed] (S) to (D);
		\draw[->] (Mp) to (M);
		\draw[->] (Np) to (N);
		\end{tikzpicture}
        \caption{A modification of the situation in \Cref{ex: doctor VoI fair} such that the causal structure is different but the observational distribution over $(S,N',M')$ is unchanged. Under this causal modification, the optimal policy under the original $\{M\}$-VoI unfair utility satisfies interventional path-specific fairness despite being intuitively unfair.}
        \label{fig: confounder}
	\end{figure}

     \begin{figure}[t]
		\centering
		\begin{tikzpicture}[node distance={20mm},thick,main/.style={draw,rectangle},scale=0.5]
		\node[fill=none,style=main](S) at (0,-2) {$S$};
		\node[fill=none,style=main](Mp) at (3,-.3) {$M'$};
		\node[fill=none,style=main](Np) at (3,-3.7) {$N'$};
		\node[fill=none,style=main](M) at (6,-.3) {$M$};
		\node[fill=none,style=main](N) at (6,-3.7) {$N$};
		\node[fill=cyan,style=main](D) at (4.5,-2) {$D$};
		\node[fill=yellow,style=main](U) at (9,-2) {$U$};
		\draw[->] (S) to (Mp);
		\draw[->] (S) to (Np);
		\draw[->] (M) to (U);
		\draw[->,dashed,DarkRed] (Np) to (D);
		\draw[->] (N) to (U);
		\draw[->] (D) to (U);
		\draw[->,dashed,DarkRed] (S) to (D);
		\draw[->] (Mp) to (M);
		\draw[->] (Np) to (N);
		\end{tikzpicture}
        \caption{A modification of the situation in \Cref{ex: doctor VoI fair} such that $M'$ is not observed. In this case, we must consider the unfair paths to be $\mathcal{M}=\{S\to D,S\to N',\to D\}$, otherwise, intuitively unfair policies would be allowed. But in this situation, it may be considered intuitively fair to use $S$ to infer $M$, which is also precluded by path-specific fairness. An $\{M\}$-VoI fair policy would use $S$ to adjust for the human bias in $N'$ and to infer $M$.}
        \label{fig: noMprime}
	\end{figure}

\section{Further details on experiments}
\subsection{Further details on \Cref{ex: doctor VoI fair continued}}\label{app: appropriate VoI-fair}

\subsubsection{
Derivation
of the corresponding VoI-fair utility function} \label{app: analysolumedicalstaff} 
Let  $\upsilon_{\bm{w}}\in \Upsilon^*$ 
with $\bm{w} = (w_1, w_2, w_3) \in \mathbb{R}^3$
be given by 
$\upsilon_{\bm{w}}: (D,S,N,M)\mapsto 
\mathbbm{1}(D=1)(
w_1S+w_2N+w_3M)$. The optimal policy in  $\Pi^{\{S,M\}}$ under utility function $\upsilon_{\bm{w}}$ is 
\begin{align*}
    \pi(S,M)=\mathbbm{1}(\mathbb{E}(w_1S+w_2M+w_3N \ | \ S,M)>0)
\end{align*}

We have that %
\begin{align*}
    \mathbb{E}&(w_1S+w_2N+w_3M \ | \ S,M)\\
    &=w_1S +w_2\theta_S^{N'}\theta_{N'}^{N} S+w_3 M.
\end{align*}

So, $\upsilon_{\bm{w}}\in \Upsilon^*$ is $\{M\}$-VoI-fair if and only if
\begin{align*}
    w_1=-w_2\theta_S^{N'}\theta_{N'}^{N}
\end{align*}

Under the policy that hires %
everyone (which maximizes the term appearing in \Cref{def: appropriate}), we can expand the expectation of the squared difference of the modified and original utility functions 
as follows:
\begin{align*}
    &\mathbb{E}(\theta_{N}^UN+\theta_{M}^UM-w_1S-w_2N-w_3M)^2\\
    &=(\theta_{M}^U-w_3)^2\left((\theta_{M'}^M\theta_S^{M'})^2+(\theta_{M'}^M)^2+1\right)\\
    &+(\theta_{N}^U-w_2)^2((\theta_{N'}^N\theta_S^{N'})^2+(\theta_{N'}^N)^2+1)\\
    &+w_1^2\\
    &+2(\theta_{M}^U-w_3)(\theta_{N}^U-w_2)(\theta_S^{N'}\theta_{N'}^N\theta_S^{M'}\theta_{M'}^M)\\
    &-2w_1(\theta_{M}^U-w_3)\theta_S^{M'}\theta_{M'}^M\\
    &-2w_1(\theta_{N}^U-w_2)\theta_S^{N'}\theta_{N'}^N
\end{align*}

Substituting $w_1 \leftarrow -w_2\theta_S^{N'}\theta_{N'}^{N}$ and minimizing with respect to $w_2$ and $w_3$ 
yields
\begin{align*}
    &w_2=\theta_N^U\\
    &w_3=\theta_M^U+\frac{\cov(N,M)\theta_N^U}{\text{var}(M)}.
\end{align*}
So
\begin{align*}
\widetilde{\upsilon}(D,S,N,M):=& 
\mathbbm{1}(D=1)
\left(-\theta_N^U\theta_S^{N'}\theta_{N'}^{N}S+\theta_N^UN
+\left(\theta_M^U+\frac{\cov(N,M)\theta_N^U}{\text{var}(M)}\right)M\right)
\end{align*}
is the unique $\upsilon$-corresponding $\{M\}$-VoI-fair utility function in $\Upsilon^*$.
\subsubsection{
Estimation of the corresponding VoI-fair utility function by rejection sampling
}\label{app: estsolmedstaff}
We can 
illustrate
the result from the previous section
with simulated data. 
To do so, we recall
that finding the $\upsilon$-corresponding $\{M\}$-VoI-fair utility function 
$\upsilon_{\bm{w}}$
in $\Upsilon^*$ corresponds to minimizing  $\mathbb{E}(\theta_{N}^UN+\theta_{M}^UM-w_1S-w_2N-w_3M)^2$ under the constraint that 
$S$ does not have VoI relative to $\{M\}$.
Let
\begin{align*}
    L_1(\bm{w}):=&\left(\beta_{\bm{w},1}^{M}-\beta_{\bm{w},1}^{(S,M)}\right)^2+\left(\beta_{\bm{w},M}^{M}-\beta_{\bm{w},M}^{(S,M)}\right)^2\\
    &+\left(\beta_{\bm{w},S}^{(S,M)}\right)^2,
\end{align*}
where 
the $\beta$ coefficients are such that
$\mathbb{E}_{\pi=1}(U_{\bm{w}}\ | \ S,M)=\beta_{\bm{w},1}^{(S,M)}+\beta_{\bm{w},S}^{(S,M)}S+\beta_{\bm{w},M}^{(S,M)}M$ and $\mathbb{E}_{\pi=1}(U_{\bm{w}}\ | \ M)=\beta_{\bm{w},1}^{M}+\beta_{\bm{w},M}^{M}M$ 
with $U_{\bm{w}} := \upsilon_{\bm{w}}(D, S, N, M)$. 
Here, the subscript $\pi=1$ indicates that the expectation is taken under the policy that hires everyone. 
$L_1(\bm{w})=0$ if and only if  $S$ does not have VoI relative to $\{M\}$. 
Let \begin{align*}
    L_2&(\bm{w}):=\mathbb{E}(\theta_{N}^UN+\theta_{M}^UM-w_1S-w_2N-w_3M)^2.
\end{align*} 
$L_2(\bm{w})$ is the 
expected squared
difference between $\upsilon_{\bm{w}}(N,M,S)$ and $\upsilon(N,M)$ under the policy that hires everyone
(which maximizes the term appearing in \Cref{def: appropriate}).
Finding the $\upsilon$-corresponding $\{M\}$-VoI-fair utility function in $\Upsilon^*$ corresponds to minimizing $L_2(\bm{w})$ subject to $L_1(\bm{w})=0$. 
Assume for 
this example
that we have $n=10\ 000$ 
observations from 
a random policy $\pi$ such that $\mathbb{P}_\pi(D=1\mid M',N',S)=0.1+0.9\Phi(M'+N')$ (where $\Phi$ is the cdf of a standard normal distribution). To estimate $L_1$ and $L_2$, we need observations from the policy that admits everyone (since $M$ and $N$ are not observed for applicants not hired). We obtain this by using rejection sampling on the observed data, yielding a subsample of the observed data that follows the distribution induced by the policy that hires everyone.
We estimate the $\upsilon$-corresponding $\{M\}$-VoI-fair utility function in $\Upsilon^*$ by minimizing the loss 
\begin{align*}
    \widehat{L}(\bm{w},K):=K\widehat{L_1}(\bm{w})+\widehat{L_2}(\bm{w}),
\end{align*}
where $\widehat{L_1}(\bm{w})$ is obtained by fitting linear regressions, $\widehat{L_2}(\bm{w})$ is the empirical mean, and $K\in \mathbb{R}$ is a constant chosen to enforce  $\widehat{L_1}(\bm{w})<\epsilon$ 
for a given $\epsilon > 0$, see \Cref{algo: appropriate}.

\begin{algorithm}
	\caption{Algorithm for estimating $\upsilon$-corresponding $\{M\}$-VoI-fair utility function}
		\textbf{Input:} $n$ observations $\left((S,N',M',D,M,N)_i\right)_{i\in \{1,\dots,n\}}$ from the observed policy ($M_i$ and $N_i$ being NA if $D_i=0$);
  $\epsilon\in \mathbb{R}$. 
	\begin{algorithmic}[1]
            \State Use rejection sampling to obtain a sample $\left((S,N',M',D,M,N)_i\right)_{i\in \{1,\dots,m\}}$ that follows the distribution induced by the policy that hires everyone.
            \State $K\leftarrow1$
            \State solution\_found $\leftarrow$ false
		\While{solution\_found$=$false}
            \State Find a $\bm{w}$ that minimizes $\widehat{L}(\bm{w},K)$
		\If{$\widehat{L_1}(\bm{w})>\epsilon$}
            \State $K\leftarrow 2K$
            \Else
            \State solution\_found $\leftarrow$ true
            \EndIf
            \EndWhile
		\State \textbf{Return} $\bm{w}$
	\end{algorithmic}
\label{algo: appropriate}
\end{algorithm}
We run the algorithm on 1000 different sets of observations and $\epsilon = 0.0001$;
 \Cref{table: appropriate 1} and \Cref{table: appropriate 2}
contain the results for two different sets of parameters.
The empirical estimates match the theoretical derivations from above.

The code to reproduce these results 
is available on GitHub: \url{https://github.com/FrederikHJ/Unfair-Utilities}. It runs on a standard laptop in less than 5 minutes.
\begin{table}[ht]
\centering
\caption{
Synthetic experiment based on 
\Cref{ex: doctor VoI fair continued}, 
see \Cref{app: estsolmedstaff}.
The goal is to 
improve an 
$\{M\}$-VoI-unfair utility function $\upsilon$
and estimate the parameters $w_1, w_2, w_3$ of 
the $\upsilon$-corresponding
$\{M\}$-VoI-fair utility function.
Here, we simulate data using the parameters %
$(\theta_{S}^{N'},\theta_{S}^{M'},\theta_{N'}^{N},\theta_{M'}^{M}, \theta_{N}^{U}, \theta_{M}^{U})=(1,2,3,4, 5, 6)$, so 
that the original utility function equals $\upsilon(D, N,M)=\mathbbm{1}(D=1)(5N+6M)$.
 We run the simulation and estimation 1000 times using $n=10 \ 000$ observations. We report the mean estimates and,  in parentheses, the empirical 5\% and 95\% quantiles.} \label{table: appropriate 1}
\begin{tabular}{@{}lccc@{}} \toprule
& $w_1$ & $w_2$ & $w_3$ \\ \midrule[0.6pt]
Analytic  & -15   & 5 & 7.48 \\ \midrule[0.1pt]
Estimate  & \begin{tabular}{@{}c@{}}
     -14.98 \\
     (-16.98,-13.03)
\end{tabular}   & \begin{tabular}{@{}c@{}}
     5.00 \\
     (-4.97,5.02)
\end{tabular} 
& \begin{tabular}{@{}c@{}}
     7.48 \\
     (7.29,7.69)
\end{tabular} \\ 
\bottomrule\end{tabular}
\end{table}
\begin{table}[ht]
\centering
\caption{
Similar 
to 
\Cref{table: appropriate 1}.
Here, we simulate data using the parameters 
$(\theta_{S}^{N'},\theta_{S}^{M'},\theta_{N'}^{N},\theta_{M'}^{M}, \theta_{N}^{U}, \theta_{M}^{U})=(6,5, 4,3, 2, 1)$, so 
that the original utility function equals $\upsilon(D, N,M)=\mathbbm{1}(D=1)(2N+M)$.
We run the simulation and estimation 1000 times using $n=10 \ 000$ observations. We report the mean estimates and,  in parentheses, the empirical 5\% and 95\% quantiles.} \label{table: appropriate 2}
\begin{tabular}{@{}lccc@{}} \toprule
& $w_1$ & $w_2$ & $w_3$ \\ \midrule[0.6pt]
Analytic  & -48   & 2 & 4.06 \\ \midrule[0.1pt]
Estimate  & \begin{tabular}{@{}c@{}}
     -47.99 \\
     (-50.43,-45.50)
\end{tabular}   & \begin{tabular}{@{}c@{}}
     2.00 \\
     (1.96,2.04)
\end{tabular} 
& \begin{tabular}{@{}c@{}}
     4.06 \\
     (3.89,4.23)
\end{tabular} \\ 
\bottomrule\end{tabular}
\end{table}
\subsection{Further details on \Cref{app: college admission}}\label{app: college example details}

\subsubsection{Data-generating process}\label{app: sim college example}
We consider the 
 following assignments for $i \in \{1, \ldots, n\}$:
\begin{align*}
    S_i&:=\varepsilon_{S_i}\sim \text{Ber}\left(\frac{2}{3}\right)\\
    S^*_i&:=\mathbbm{1}(S_i=1)-\mathbbm{1}(S_i=0)\\
    E_i&:=\varepsilon_{E_i}\sim \calN(0,1)\\
    \varepsilon_{T_i} &\sim \calN(0.3,1)\\
    T_i&:=E_i+\varepsilon_{T_i}S^*_i\\
    \varepsilon_{R_i}&\sim \calN(1,1)\\
    R_i&:=E_i+\varepsilon_{R_i}S_i^*\\
    \varepsilon_{Y^*_i}&\sim \calN(0,2)\\
    Y^*_i&:=E_i+0.5R_i+0.5T_i+\varepsilon_{Y^*_i}\\
    Y_i&:=\mathbbm{1}(Y^*_i>\widehat{F^{-1}_{Y^*}}(0.7)),
\end{align*}
where $\widehat{F^{-1}_{Y^*}}$ is the `empirical' quantile function based on $Y^*_1, \ldots, Y_n^*$, and $S^*_i$ is a transformation of $S_i$ such that $S^*_i\in \{-1,1\}$, where $E_i$ are variables introducing additional dependence between the variables. 

\subsubsection{Approximation of $w$}\label{app: estimate w}
This appendix shows how to approximate $w$ given knowledge about the data-generating process. In practice, this derivation may not be feasible because, for example, we may not have access to data from the policy that admits every applicant.
We want to find $w$ such that $S$ does not have VoI relative to $\{T\}$ under utility function $\widetilde{U}_w :=\widetilde{\upsilon}_w(D,Y,S) :=D^\top(Y+wS)$.
We simulate a training set and a test set ($n=100 \ 000$ for each) from the policy that admits everyone, $\pi(S,T,R)=1$.
Based on the training data, we fit logistic models to estimate $\mathbb{E}(Y_i \ | \ T_i)$, $\mathbb{E}(Y_i \ | \ S_i,T_i)$, 
$\mathbb{E}(S_i \ | \ T_i)$, and $\mathbb{E}(Y_i \ | \ S_i,T_i,R_i)$. 
The optimal policy in $\Pi^{\{T\}}$ under utility function $\upsilon_w$ is the policy that admits applicant $j$ 
if and only if
$\mathbb{E}(Y_j  \ | \ T_j)+w\mathbb{E}(S_j \ | \ T_j)$ is larger than the median of $\left(\mathbb{E}(Y_i \ | \ T_i)+w\mathbb{E}(S_i \ | \ T_i)\right)_{i\in \{1,\dots, n\}}$. Similarly, the optimal policy in  $\Pi^{\{S,T\}}$ under utility function $\upsilon_w$ is a policy that admits applicant $j$ 
if and only if $\mathbb{E}(Y_j \ | \ S_j,T_j)+wS_j$ is larger than the median of $\left(\mathbb{E}(Y_i \ | \ S_j, T_i)+wS_i\right)_{i\in \{1,\dots,n\}}$. So we can use the estimated expected values to %
approximate
the optimal policies $\pi_w^{\{T\}}$ and $\pi_w^{\{S,T\}}$ in $\Pi^{\{T\}}$ and $\Pi^{\{S,T\}}$ under utility function $\upsilon_w$, respectively. On the test data, we estimate $\mathbb{E}_{\pi_w^{\{T\}}}(\widetilde{U_w})$ and $\mathbb{E}_{\pi_w^{\{S,T\}}}(\widetilde{U_w})$ and 
choose (using automatic differentiation)
$w$ such that $\mathbb{E}_{\pi_w^{\{S,T\}}}(\widetilde{U_w})=\mathbb{E}_{\pi_w^{\{T\}}}(\widetilde{U_w})$.

\subsubsection{Multiple realizations of simulations}\label{app: multiple realizations}
We run the entire simulations and estimations of $\omega$ 100 times. We report averages and empirical 5\% and 95\% quantiles rounded to the nearest whole number: On average, 76 (31--113) fewer persons end up graduating after modifying the utility function, 2089 (1872--2346) more persons from the minority group are admitted, and 426 (376--490) more persons from the minority group graduate. 
The code to reproduce these results 
is available on GitHub: \url{https://github.com/FrederikHJ/Unfair-Utilities}. It runs on a standard laptop in less than 5 minutes.

\subsection{Further details on \Cref{ex: compas}}
\subsubsection{Details on XGBoost}\label{app: compas}
We run XGBoost in Python using standard parameters, maximum tree depth at 3, and learning rate 0.1 \citep{chen2016xgboost}. We specify the loss function ourselves by providing gradient and Hessian. Since XGBoost does not give predictions in $(0,1)$ per default, we use standard logistic reparametrization $D_i=\frac{e^{\eta_i}
}{e^{\eta_i}+1}$. We train XGBoost to return $(\eta_1,\dots, \eta_N)\in \mathbb{R}^N$ as to minimize the loss \begin{align*}
       \sum_{i=1}^N\left(\frac{e^{\eta_i}
}{e^{\eta_i}+1}-Y_i\right)^2+\lambda\cdot \left(\frac{e^{\eta_1}
}{e^{\eta_1}+1},\dots, \frac{e^{\eta_N}
}{e^{\eta_N}+1}\right)\bm{\mathbbm{1}}(\text{African-American}),
\end{align*} 
where $\bm{\mathbbm{1}}(\text{African-American})\in \{0,1\}^N$.

\newpage
 
\subsubsection{Histogram of prediction in \Cref{ex: compas}}\label{app: histograms}
Figure~\ref{fig: hist-compas} shows the predictions of the original and modified utilities on the third fold.

     \begin{figure}[ht!]
        \centering
        \includesvg[width=0.7\textwidth]{original.svg} %
        \vspace{1em} %
        \includesvg[width=0.7\textwidth]{modified.svg} %
        \caption{Histograms for how the predictions made by the corresponding optimal policy (on the third fold) change when we modify the 
        original utility function (top) to be $\{\text{charge degree}\}$-VoI-fair (bottom) (both policies are based on $\rO^D$). The plot uses data from the first of the 20 runs.}
        \label{fig: hist-compas}
	\end{figure}

\section{Proofs} 
\label{app: proof graphical}
\subsection{Proof of Proposition~\ref{prop: graphical}}
It is shown in \citep{Everitt2021} that a DAG $\mathcal{G}$ over $(\rX, D, U)$, where $U$ is a descendant of $D$ and $U$ has no children, admits VoI for $S\in \rX \backslash \DE^D$  relative to $\rM\subseteq \rX \backslash (\DE^D\cup \{S\})$ if and only if
\begin{align*}
S  \not \perp_{\mathcal{G}_{\PA^D:=\rM \cup \{S\}}} U \ | \ \rM \cup \{D\}.
\end{align*}
Since we assume that $S\notin \rM$ this simplifies to the following d-separation. 

\newtheorem*{proprestate}{\Cref{prop: graphical}}
\begin{proprestate}

 \Paste{propgraphical1}
 \begin{equation*}
      {S  \not \perp_{\calG_{\PA^D:=\rM}} U \mid \rM.}
 \end{equation*} 
 \Paste{propgraphical2}
\end{proprestate}

\begin{proof}
	Assume that $\mathcal{G}$ admits VoI for $S$ relative to $\rM$. Then, 
	\[S  \not \perp_{\mathcal{G}_{\PA^D:=\rM\cup \{S\}}} U \ | \  \rM \cup \{D\}.\]
	Since in this statement, we condition on $\rM=\PA^D\backslash \{S\}$ and $D$, an open path cannot go through the edge $S\to D$. Thus, there 
 also exist an open path in $\mathcal{G}_{\PA^D:=\rM}$:
	\[S  \not \perp_{\mathcal{G}_{\PA^D:=\rM}} U \ | \ \rM \cup \{D\}.\] 
	If there is an open path between $S$ and $U$ given $\rM\cup \{D\}$, then this path is also open given $\rM$. The only ways an open path could be blocked by no longer conditioning on $D$ are 
	\begin{adjustwidth}{1cm}{1cm}
		(1) if $D$ is a collider relative to the path, or\\
		(2) if the path goes through a collider that is an ancestor of $D$. 
	\end{adjustwidth}
	(1) is impossible as we also condition on $\rM=\PA^D$. 
 If (2) occurs, then either the collider is in $\rM=\PA^D$ and the path is still open, or the collider is an ancestor of a node in $\rM=\PA^D$, which also implies that the path is still open. So 
	\[S  \not\perp_{\mathcal{G}_{\PA^D:=\rM}} U \ | \ \rM. \]
	
	Assume now that %
 $S \not\perp_{\mathcal{G}_{\PA^D:=\rM}} U \ | \  \rM$. An open path between $S$ and $U$ given $\rM=\PA^D$ cannot go through $D$.  Therefore, this path is also open given $\rM \cup \{D\}$:
	\[S  \not \perp_{\mathcal{G}_{\PA^D:=\rM}} U \ | \ \rM \cup \{D\}.\]	
	Finally, adding an edge $S\to D$ does not block any paths.  
\end{proof}

\end{document}